\documentclass{ieeeaccess}
\usepackage{cite}
\usepackage{amsmath,amssymb,amsfonts}
\usepackage{algorithmicx}
\usepackage{textcomp}

\usepackage{xcolor}
\usepackage{ifpdf}
\usepackage{color}

\usepackage{soul}
\usepackage{lineno,hyperref}
\modulolinenumbers[5]
\usepackage{booktabs}
\usepackage{bm}
\usepackage{multirow}
\usepackage{array}
\usepackage{graphicx}
\usepackage{algpseudocode}
\usepackage{algorithm}
\usepackage{mathtools}
\usepackage{balance}
\usepackage{subcaption}
\usepackage[justification=centering]{caption}
\usepackage{blindtext}
\usepackage{balance}
\def\BibTeX{{\rm B\kern-.05em{\sc i\kern-.025em b}\kern-.08em
		T\kern-.1667em\lower.7ex\hbox{E}\kern-.125emX}}
	
\begin{document}

	\doi{}
	
	\title{Hierarchical Convolutional Neural Network with Feature Preservation and Autotuned Thresholding for Crack Detection}
\author{\uppercase{{Qiuchen Zhu}\authorrefmark{1},
		\uppercase{Tran Hiep Dinh}\authorrefmark{2,3}, \uppercase{Manh Duong Phung}}\authorrefmark{1,3},
	and \uppercase{Quang Phuc Ha}\authorrefmark{1}} 
	\address[1]{School of Electrical and Data Engineering,
		University of Technology Sydney, Sydney, NSW 2000, Australia (e-mail: Qiuchen.Zhu@uts.edu.au; ManhDuong.Phung@uts.edu.au; Quang.Ha@uts.edu.au)}
	\address[2]{UTS - VNU Joint Technology and Innovation Research Centre (JTIRC), Hanoi, Vietnam (e-mail: tranhiep.dinh@vnu.edu.vn)}
\address[3]{University of Engineering and Technology, Vietnam National University, Hanoi, Vietnam }

	
	\markboth
	{Zhu \headeretal: Hierarchical Convolutional Neural Network with Feature Preservation and Autotuned Thresholding for Crack Detection}
	{Zhu \headeretal: Hierarchical Convolutional Neural Network with Feature Preservation and Autotuned Thresholding for Crack Detection}
	
	\corresp{Corresponding author: Qiuchen Zhu (e-mail: Qiuchen.Zhu@uts.edu.au).}

\begin{abstract}
	Drone imagery is increasingly used in automated inspection for infrastructure surface defects, especially in hazardous or unreachable environments. In machine vision, the key to crack detection rests with robust and accurate algorithms for image processing. To this end, this paper proposes a deep learning approach using hierarchical convolutional neural networks with feature preservation (HCNNFP) and an intercontrast iterative thresholding algorithm for image binarization. First, a set of branch networks is proposed, wherein the output of previous convolutional blocks is half-sizedly concatenated to the current ones to reduce the obscuration in the down-sampling stage taking into account the overall information loss. Next, to extract the feature map generated from the enhanced HCNN, a binary contrast-based autotuned thresholding (CBAT) approach is developed at the post-processing step, where patterns of interest are clustered within the probability map of the identified features. The proposed technique is then applied to identify surface cracks on the surface of roads, bridges or pavements. An extensive comparison with existing techniques is conducted on various datasets and subject to a number of evaluation criteria including the average F-measure ($AF_\beta$) introduced here for dynamic quantification of the performance. Experiments on crack images, including those captured by unmanned aerial vehicles inspecting a monorail bridge. The proposed technique outperforms the existing methods on various tested datasets especially for GAPs dataset with an increase of about 1.4\% in terms of $AF_\beta$ while the mean percentage error drops by 2.2\%. Such performance demonstrates the merits of the proposed HCNNFP architecture for surface defect inspection.
	
\begin{keywords}
deep learning, crack detection, hierarchical convolutional neural network, feature preserving, thresholding.
\end{keywords}
\end{abstract}

\titlepgskip=-15pt

\maketitle

\section{Introduction}
\label{intro}
Surface inspection plays an important role in the health surveillance and hazard control of roads, bridges, pavements or tunnels. Effective maintenance and damage prevention of transport infrastructure rely on prompt detection for defects in transportation infrastructure such as cracks, edge failures, potholes, rutting, subsidence, or any surface deterioration \cite{cao2020review}. For this, the inspection conducted manually by professional practitioners, wherein dangerous and unattainable sites would limit the effectiveness of human inspection. With advances in unmanned aerial vehicles (UAVs) and field robotics, machine vision-based systems are introduced to fulfill those inspection tasks \cite{song2019real}. For automatic inspection, successful identification of defect features on infrastructure surface requires the development of feasible, robust and effective detection algorithms.

In visual inspection from captured images, an intensity shift indicates a contrast between defective spots and their surrounding pixels in the color space. Based on the referred source of information, various methods for crack detection have been proposed. Initially, thresholding approaches are employed to execute fast detection by solely exploiting the statistics of intensity. Based on thresholding, early trials for surface imperfectness detection have been conducted with hybrid utilization of intensity and geometrics \cite{li2018automatic}. Other methods roughly classify the image according to the distribution of pixel intensity \cite{dinh2019summit}. To alleviate the interference from noisy textures, a scanning kernel like the Gabor filter has been developed in the frequency domain \cite{hoang2019novel}. This scanning kernel can be used for pixel-wise extraction of the local geometric information about crack pixels and sharing the similarity to current convolutional kernels. To judge on surface defect, a probability model is usually formulated to determine the presence of cracks. An entropy formulation is introduced to guide the pavement crack detection based on saliency and statistical features \cite{xu2013pavement}. Alternatively,  Minimal Path Selection (MPS) on a single scale image \cite{kaul2011detecting} or multi-scale fusion \cite{li2018automatic} can be used in searching crack seeds. Promising results have suggested the formation of a trainable framework with improved robustness for crack detection. 

With the increasing interest in artificial intelligence, machine learning (ML) approaches are introduced for heuristic abnormality detection. To this end, several ML techniques using shallow linear regression models like support vector machine (SVM) \cite{ai2018auto} and random forests \cite{shi2016automatic} have been applied for crack detection. Such learning approaches provide an adaptive solution in disposing of a variety of crack patterns. However, the prediction accuracy may be limited by the model simplification and the available computational capacity in practice. On the other hand, deep convolutional neural networks (DCNN) \cite{zhang2018application} have been proposed as a probabilistic learning framework with a modest processing time that becomes very attractive for real-time applications. This technique provides an effective solution to challenges in semantic segmentation \cite{badrinarayanan2017segnet} owing to the capability of multiple-level abstraction and deep breakdown for identified features. Such promising results suggest the use of DCNN to identify a surface defect with binary segmentation.

In deep learning, often, the network is sequentially structured and finalized by fully-connected layers. Such structures may be computationally ineffective and cause blurred representation \cite{Zhu2019}, leading to a drop in the prediction accuracy. Recently, the emerging hierarchical structure \cite{wang2020convolutional} has been applied to deal with the blurry problem. The potential of this framework in crack detection has been verified in \cite{zou2018deepcrack}. In addition, a well-designed filter can be incorporated in the hierarchical convolutional learning process to extract the probability map of features \cite{Zhu2019}. Accordingly, it is promising to seek a suitable post-processing approach that can offer a more effective determination of crack and background with hierarchical DCNN. 

Here, a hybrid approach is proposed, integrating a hierarchical convolutional neural network with feature preserving and the contrast-based autotuned thresholding (CBAT) technique to identify surface cracks of roads, bridges or pavements via aerial photography, obtained by a formation of unmanned aerial vehicles (UAV) \cite{phung2019system}. The collected images are processed by the proposed neural network first for a probability map of a potential defect and then its features are further extracted by CBAT binarization for identification. Experiments on different datasets from \cite{shi2016automatic} and on images captured by our UAVs during the inspection of a monorail bridge indicate the advantage of the hierarchical convolutional neural network with feature preservation (HCNNFP) proposed in comparison with some crack detection approaches available in the existing literature. Various frameworks for crack detection and specific datasets are considered in a number of experiments for comprehensive assessment on the merits of our HCNNFP as well as its improvement over other post-processing methods. 

The paper is organized as follows. After the introduction, Section \ref{relevant} discusses the relevant work for deep-learning-based crack detection. Section \ref{networks} describes the architecture of the proposed framework for crack detection. Section \ref{postprocessing} presents our thresholding technique for post-processing. Section \ref{experiment} introduces our UAV system for image capturing, the datasets, the setup of two experiments respectively for comparison with relevant deep learning techniques and for post-processing with binarization. Section \ref{discussion} demonstrates the experimental results along with their discussion. Finally, a conclusion is drawn in Section \ref{conclusion}.  

\section{Relevant work}\label{relevant}
In this section, key technologies using convolutional neural networks (CNN) in crack detection tasks are briefly discussed. 
Judging on a pipeline structure, CNN methods can be divided into sequential and hierarchical models. In sequential models, only the final output is involved in benchmark matching. For hierarchical models, features from multiple processing branches and the ground-truth can be utilized to collectively contribute to improving the fitness of the detection result. 

Current CNN models for crack detection are listed in Table \ref{sum}, showing also the various methods that have been used to enhance the extracted features of the crack pattern. The sequential models include basic CNN \cite{zhang2016road}, CNN with metaheuristics (CNN-M) \cite{nhat2018automatic}, deep fully CNN (FCN) \cite{dung2019autonomous}, Cracknet \cite{zhang2017automated}, Cracknet-V \cite{fei2019pixel}, and densely-connected CNN \cite{mei2020densely}. Those hierarchical CNN models relevant to this work are DeepCrack \cite{zou2018deepcrack}, feature pyramid and hierarchical boosting network (FPHBN) \cite{yang2019feature}, U-Net\cite{liu2019computer}, CNN with na\"{\i}ve Bayesian data fusion (NB-CNN) \cite{chen2017nb}, weakly-supervised DCNN (WS-ConvNet) \cite{chen2020robust}, PGA-Net \cite{dong2019pga} and SDD-Net \cite{choi2019sddnet}. Methods used for feature enhancement include deconvolutional decoders (D), residual modules (RM), probabilistic representation (PR), and statistic post-processing (SP). In the encoder-decoder structure, the crack patterns are to be rescaled with key indices recorded. With the preservation of those coordinates, the detailed patterns of crack features can be reasonably refilled with deconvolutional decoding. In RM, a combination of the original and processed features can be used to provide a residual effect like with human eyes, i.e., remembering the silhouette of the object that has previously been observed. The following step is to compensate for missing patterns using this residual effect. Alternatively, the feature maps are converted into a representation in the probabilistic space PR. In this case, every pixel will be assigned a possibility score in the range of $(0,1)$ to evaluate the likelihood of a crack. As a result, the prediction of the model becomes less overconfident on uncertainty and mislabels, contributing positively to the reduction of false-positive rate. The last approach SP using a global optimizer with statistic post-processing tools to extract the result from CNN. The accuracy of the detection can be improved by filtering out outliner labels with a total threshold or some verification mechanism. 

\begin{table}[tbh!]
	\centering
		\renewcommand\thetable{I}
	\begin{tabular}{l|l|llll}
		\hline
		Models                                                                                           & \begin{tabular}[c]{@{}l@{}}Type of \\ architecture\end{tabular} & \begin{tabular}[c]{@{}l@{}}D\end{tabular} & \begin{tabular}[c]{@{}l@{}} RM\end{tabular} & \begin{tabular}[c]{@{}l@{}}PR\end{tabular} & \begin{tabular}[c]{@{}l@{}} SP \end{tabular} \\ \hline
		Basic CNN \cite{zhang2016road}                   & Sequential                                                      &                                                                  &                                                             &                                                                         &                                                                      \\
		CNN-M \cite{nhat2018automatic}                                                & Sequential                                                      &                                                                  &                                                             &                                                                         & \checkmark                                                 \\
		FCN \cite{dung2019autonomous}                                                   & Sequential                                                      & \checkmark                                             &                                                             &                                                                         &                                                                      \\
		CrackNet \cite{zhang2017automated}                                              & Sequential                                                      &                                                                  &                                                             &                                                                         &                                                                      \\
		CrackNet-V \cite{fei2019pixel}                                                  & Sequential                                                      &                                                                  &                                                             &                                                                         &                                      \checkmark                                  \\	
		Densely connected CNN \cite{mei2020densely}                    & Sequential                                                      &                                                                  & \checkmark                                        &                                                                         & \checkmark                                                 \\
		DeepCrack \cite{zou2018deepcrack}                                               & Hierarchical                                                    & \checkmark                                             & \checkmark                                        & \checkmark                                                    &                                                                      \\
		FPHBN \cite{yang2019feature} & Hierarchical                                                    & \checkmark                                             & \checkmark                                        & \checkmark                                                    &                                                                      \\
		U-Net \cite{liu2019computer}                                    & Hierarchical                                                    &    \checkmark                                                                &                                                             &                                                                         &                                                 \\
		NB - CNN \cite{chen2017nb}                                             & Hierarchical                                                    &                                                                  &                                                             & \checkmark                                                    &                                                                      \\

		WS-ConvNet \cite{chen2020robust}                        & Hierarchical                                                    &                                                                  & \checkmark                                        &                                                                         & \checkmark                                                 \\  
		
		PGA-Net \cite{dong2019pga}                                             &  Hierarchical                                                    &    \checkmark                                                               &    \checkmark                                                          &                                                    &                                                                      \\
		SDD-Net \cite{choi2019sddnet}                                             &  Sequential                                                    &                                                                   &    \checkmark                                                          &                                                    &                                                                      \\	\hline
	\end{tabular}
	\caption{Summary of CNN models with different architecture and different feature enhancement methods.}
	\label{sum}
\end{table}

In this paper, the above enhancement methods are integrated to create a new processing pipeline for crack detection.  The contributions of this paper can be summarized as follows: 
\begin{itemize}
	\item Among hierarchical architectures, the HCNNFP network proposed in this paper is different from the Deepcrack by a feature preserving branch. As such, it is more comprehensive in surface crack detection by using the combination of geometrical and statistic information, whereby feature abstraction is enhanced by an additional side branch in the encoder to reduce estimation error caused by redundant nonlinearity. 
	\item An iterative approach is proposed to automatically search for an optimized threshold of the probability map for features generated from the proposed DCNN, and as a result, to avoid the time consumption in the search while increasing the accuracy of generated feature maps. 
	\item A dynamic measure to evaluate the fitness of defect detection, assessing the average performance under a range of weights in conjunction with the commonly-used F-measure using a single pre-determined weight.
\end{itemize}	
	\section{Hierarchical Convolutional Neural NETWORK WITH FEATURE PRESERVATION}\label{networks}
	In this section, a novel DCNN approach called the hierarchical convolutional neural network with feature preservation (HCNNFP) is proposed to obtain a probability map of surface defects from the input image. Here, unlike the original hierarchical CNN, a feature preserving branch is augmented to adjust the weights of the abstraction from upper-layers, and hence, resolving the nonlinearity trade-off to improve the network performance. 
	
	\subsection{Convolutional neural networks}
	Our detection method is based on the inference in a convolutional neural network (CNN).
	To formulate the classification problem, let us first define a training sample as $D = \{(X,Y)\}$, where $X=\left\{x_{ij}| i,j \in (I \times J)\right\}$ and $Y=\left\{y_{ij}| i,j \in (I \times J)\right\}$ respectively represent the pixel values of the original image of size $I \times J$ and its corresponding annotated mask of cracks, both containing $I \times J$ pixels. In the context of defect identification, the ground-truth mask $y_{ij}$ can take a binary value determined as,
	\begin{equation}
	y_{ij} = \left\{
	\begin{array}{lr}
	1, ~ \ x_{ij}  \text{\ - abnormal pixel in the mask},\\
	0 ~  \text{\ otherwise}.
	\end{array}
	\right.
	\end{equation}
	In a network model, the judgment on crack candidates is made from a layer-by-layer inference.
	Such inference is deduced from the basic structure of multilayer perceptrons (MLP) \cite{wang2016towards}.  Suppose we have a $L$-layer MLP to predict the possibility map of defect candidates. For crack detection, the learning process targets at the best fitness to the annotated crack map. For an input matrix $X^l$ in $l$-th layer, the learning process is an optimization problem formulated as
	\begin{equation}\begin{array}{c}
	\min _{\left\{{W}^{l}\right\},\left\{{b}^{l}\right\}}\left\|P({X}^{L})-{Y}\right\|_{F}+\lambda \sum_{l}\left\|{W}^{l}\right\|_{F}^{2} \\
	\text { subject to } {X}^{l}=a\left({X}^{l-1} {W}^{l}+{b}^{l}\right), l=1, \ldots, L-1 \\
	{X}^{L}={X}^{L-1} {W}^{L}+{b}^{L},
	\end{array}\end{equation}
	where $W^l$ and $b^l$ represent the weights and bias at the $l$-th layer, $a(\cdot)$ is the nonlinear activation function, $\lambda$ is the coefficient for controlling the scale of regularization, and $P(\cdot)$ is an arbitrary statistic function to express the possibility of crack candidates.

 While CNN shares a similar architecture with that of MLP, it introduces the convolution to biologically stimulate the visual perception of cortex cells within a receptive field. By using the convolution operation ($*$), the feature map of CNN can be expressed as a set of features:
	\begin{equation}{f}_{i j}^{l}= a \left(\left({W}^{l} * {X}\right)_{i j}+b^{l}\right).\end{equation}

	Then the target possibility $P(f_{ij})$ stimulates the conditional distribution $p(y_{ij}|x_{ij})$, the real possibility to indicate the confidence level of a pixel looking like a crack.	However, it would be impractical for the neurons to be fully connected pixel-wisely due to the exponential increase of computational load. In this application, the non-zero weights are actually restrained within a certain kernel to present the logistic relation between a pixel and its limited neighborhood. The smallest kernel size that can provide the abstraction of the image is $3 \times 3$ (a $1 \times 1$ kernel can just output the original information). For a larger receptive field, the size of the kernel can be extended appropriately. Within the receptive field, the values of neighboring pixels collectively determine the intensity of a particular pixel on the feature map. As a result, the output can be a highly dependent abstraction of the crack patterns.
	
	\subsection{Network structure}
\begin{figure*}[h]
		\centering
		\includegraphics[width=0.75\textwidth]{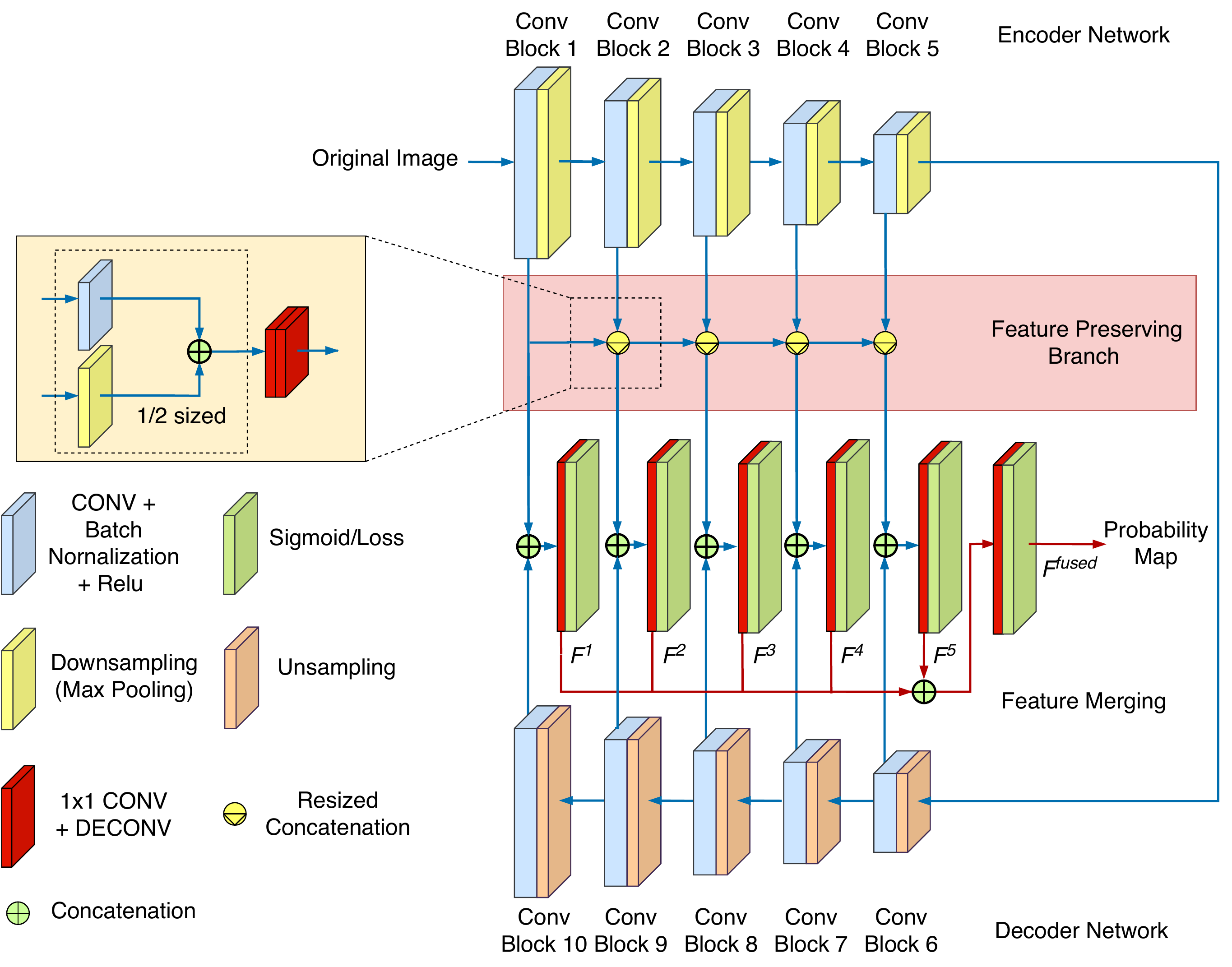}
		\caption{HCNNFP- network architecture}
		\label{fig:net}
	\end{figure*}

	 The proposed symmetric network architecture is shown in Fig. \ref{fig:net}, based on the framework of DeepCrack\cite{zou2018deepcrack} with the same structural parameters such as the kernel size and the number of channels. The encoder consists of 5 convolutional blocks with 13 convolutional layers. Each convolutional layer contains a standard size-invariant convolution operation for the abstraction of crack features. With the combination of several convolutional layers, each feature map of the current scale is created at the end of each block. For downsampling, a max-pooling layer after each block compresses the image into quarter-size, preserving the values and indices of the local maxima. The shrinkage of the image consequently yields an increase in the size of the receptive field (RF) for the following layer, leading to a sparser feature map in the next block. 
	
In the first two blocks, the convolutional kernels are relatively small compared to the size of the input. Accordingly, the first two blocks mainly preserve the detailed features of the original image. However, unlike the early CNNs aiming at small-size images as $28 \times 28$ in the Modified NIST dataset \cite{lecun1998gradient}, the current deep learning approaches are designed to segment images with a size of at least hundreds by hundreds. Hence, the basic kernel in $3 \times 3$ is insufficient for abstraction, and as such, two convolutional layers are combined to stimulate the receptive field of a $5 \times 5$ kernel but with fewer parameters involved. The outputs from the third block are equal or less than the 1/16 size of the original image. Therefore, the receptive field of those blocks should be extended for preserving features. Consequently, an additional convolutional layer is added in each of those blocks. Here, a feature preserving branch is augmented at the output of the convolutional blocks in the encoder to adjust the level of abstraction from upper-layers by concatenation with the downsampling layer. Here, unlike the pipeline of the DeepCrack \cite{zou2018deepcrack}, which inherits the auto-encoder structure of SegNet \cite{badrinarayanan2017segnet}, the proposed preserving branch is to maintain the image features by alleviating nonlinear redundancy, as explained at the end of this Section.
	
	In our network, the decoder mirrors the structure of the encoder, with five corresponding upsampling layers to symmetrically retrieve the size of the image via the reference of saved indices. The sparse image generated from the last upsampling is refilled and reconstructed in the next blocks. With the index propagation throughout the entire pipeline, the network can restore key information of boundaries on the original image. 
	
	\subsection{Information loss} 
	Since the task of identifying a surface crack on a bridge, road or pavement can be rendered to a binary segmentation problem with two semantic groups, abnormal and intact features, the labeling error in the prediction can be evaluated by a binary entropy loss function. For the computation of a measure for information loss, let us define  $F^k=\{f^k_{ij}|k=1,\dots,5\}$ as the feature map under the zooming scale $k$ and $F^{fused}=\{f^{fused}_{ij}\}$ as the fused map accordingly. The red modules depicted in Fig. \ref{fig:net} illustrates the formation of those feature maps. 
	For an arbitrary feature $f_{ij}$, its pixel-wise probability $P(f_{ij})$ can be expressed by a sigmoidal function as,
	\begin{equation}
	P(f_{ij}) = \frac{1}{1+e^{-f_{ij}}}.
	\label{eq.sig}
	\end{equation} 
	
	In terms of binary entropy, a pixel-wise feature at a convolutional block is either abnormal or intact. By considering it as a random variable, the associated information loss for feature $f_{ij}^{k}$ at the $k^{th}$ convolutional block can be expressed via its entropy as,
	\begin{equation}
	l(f_{ij}^{k}) =-y_{ij}ln(P(f_{ij}^{k}))-(1-y_{ij})\ln(1-P(f_{ij}^{k})).
	\label{eq:loss}
	\end{equation}
	
	Since the ground-truth mask contains only logical values 0 and 1,  the information loss or entropy of Eq. (\ref{eq:loss}) is rewritten as,
	\begin{equation}
	l(f_{ij}^{k}) = \left\{
	\begin{array}{lr}
	-\ln P(f_{ij}^{k}), &\   \ y_{ij}=1\\
	-\ln (1-P(f_{ij}^{k})), &\   \ y_{ij}=0. 
	\end{array}
	\right.
	\end{equation}
	The accuracy of the prediction relies on the fitness of every feature map in comparison with the ground-truth mask. Hence, all the corresponding probability maps are responsible for the loss function, including for all fused pixels and all convolutional blocks. Accordingly, the total loss $\mathcal{L}$ of an image should represent the superposition of the pixel-wise losses for each convolutional block for all feature maps and the fused map for all pixels, i.e.
	\begin{equation}
	\mathcal{L}=\sum_{i=1}^{I}\sum_{j=1}^{J}\left(l(f_{ij}^{fused})+\sum_{k=1}^{5} l(f_{ij}^{k})\right).
	\end{equation}
	
	\subsection{Performance enhancement}
	For U-shape networks like the U-net \cite{liu2019computer}, enhancement of features can be achieved with a comprehensive design to avoid ambiguously stacking additional channels. Here, a feasible structure is implemented with an alternative allocation inside the network. To analyze the performance improvement of the proposed HCNNFP, we consider the nonlinear nature of the network and then adjust its structure to simultaneously reduce nonlinearity while preserving the image features by resolving the trade-off between them.
	
	\subsubsection{Feature preservation versus nonlinearity}	From the probabilistic view, the attribution of a pixel can be properly described by either an abnormal or intact pixel member corresponding to two random events $EV_1$ and $EV_0$, respectively. Specifically, $EV_1$ is the event that the sampled pixel belongs to the abnormal group and $EV_0$ is when it belongs to the intact area.  Accordingly, given an observation on the pixel $x_{ij}$, two conditional probabilities are defined, namely the probability $P\left(EV_{1}| x_{ij}\right)$ or $P\left(EV_{0}| x_{ij}\right)$ that $x_{ij}$ belongs to surface abnormality such as a crack or not. To identify a potential defect, let us consider the probability $P\left(EV_{1}| x_{ij}\right)$. From Bayes's rule, the conditional probability of a crack on a road, pavement or bridge, given pixel $x_{ij}$ is expressed as $$P\left(EV_{1}| x_{ij}\right) =  \frac{P\left(EV_{1}, x_{ij}\right)}{P(x_{ij})},$$ or 
	\begin{equation}
	\begin{aligned}
	P\left(EV_{1}| x_{ij}\right) &=\frac{P\left(x_{ij} | EV_{1}\right) P\left(EV_{1}\right)}{P\left(x_{ij} | EV_{1}\right) P\left(EV_{1}\right)+P\left(x_{ij} | EV_{0}\right) P\left(EV_{0}\right)}\\&=\frac{1}{1+\frac{P\left(x_{ij} | EV_{0}\right) P\left(EV_{0}\right)}{P\left(x_{ij} | EV_{1}\right) P\left(EV_{1}\right)}}\\&=\frac{1}{1+e^{-a(x_{ij})}},
	\end{aligned}
	\end{equation}

	where 
	\begin{equation}
	a(x_{ij}) = \operatorname {ln} \frac{P\left(x_{ij} | EV_{1}\right) P\left(EV_{1}\right)}{P\left(x_{ij} | EV_{0}\right) P\left(EV_{0}\right)}.
	\label{Eq.a}
	\end{equation}
	Now, it is assumed that those conditional probabilities follow the Gaussian process $\mathcal{N}(\mu_{0,1}, \mathbf{\sigma^{2}})$ with the same variance $\sigma$ \cite{Murphy:Machine} and means $\mu_{1}$ and $\mu_{0}$, respectively for the two abnormal and intact cases, we have:  
	\begin{equation}
	\begin{split}
	P(x_{ij}|EV_{0,1}) =  \frac{1}{\mathbf{\sigma}\sqrt{2\pi}} \exp \left[-\frac{(x_{ij}-\mu_{0,1})^{2}}{2\mathbf{\sigma}^{2}}\right],
	\end{split} 
	\label{Eq.gau}
	\end{equation}
	in association with the random events $EV_{0}$ and $EV_{1}$.
	By substituting Eq. (\ref{Eq.gau}) into Eq. (\ref{Eq.a}), the exponent $a(x_i)$ can be explicitly derived in the following form:
	\begin{equation}
	\begin{aligned} 
	a(x_{ij}) &= \operatorname{ln} P\left(x_{ij} | EV_{1}\right)-\operatorname{ln} P\left(x_{ij} | EV_{0}\right)+\operatorname{ln} \frac{P\left(EV_{1}\right)}{P\left(EV_{0}\right)} \\ &= \frac{\mu_{1}-\mu_{0}}{\sigma^{2}} x_{ij}+\frac{\mu_{0}^2-\mu_{1}^{2}}{2\sigma^{2}} +\operatorname{ln} \frac{P\left(EV_{1}\right)}{P\left(EV_{0}\right)} \\ &=w x_{ij}+w_{0}, \end{aligned}
	\label{eq.linear}
	\end{equation}  
	where $w= \frac{\mu_{1}-\mu_{0}}{\sigma^{2}}$ and $w_{0}=\frac{\mu_{0}^2-\mu_{1}^2}{2\sigma^{2}} +\operatorname{ln} \frac{P\left(EV_{1}\right)}{P\left(EV_{0}\right)}.$\\
	
	Therefore, by comparing Eq. (\ref{eq.sig}) and Eq. (\ref{eq.linear}), elements $f_{ij}$ of an abnormal feature as of a crack can be expressed as,
	\begin{equation}
	f_{ij} = wx_{ij} + w_{0}, 
	\end{equation}
	which is linearly-dependent on the pixel values $x_{ij}$ of the image captured on the surface under monitoring.

	In a deep learning CNN framework, it is known that the hidden layers contain inevitable nonlinearity to facilitate the information processing capacity of the network. However,  when using the sigmoidal function for probabilistic representation, the linear dependence  of image features on the input appears to limit the amount of feature information throughout the processing. This requires a compromise between feature preserving and information handling. Due to the nonlinear activation employed in each convolutional layer, the overall nonlinearity accumulates largely in the forward propagation. Therefore, the outputs from the deeper encoder network tend to depart further from the linear hypothesis, unfavorably influencing the accuracy of the probability maps as per Eqs. (8-12). Hence, some measures of compensation for nonlinearity is required to balance the trade-off between nonlinearity and the network capacity of information processing. This motivated us to develop a feature preserving branch(FPB) for the network architecture, for which a rationale is given in the following.
	
	\subsubsection{Feature preserving branch}
\begin{figure}[h]
		\centering
		\includegraphics[width=0.4\textwidth]{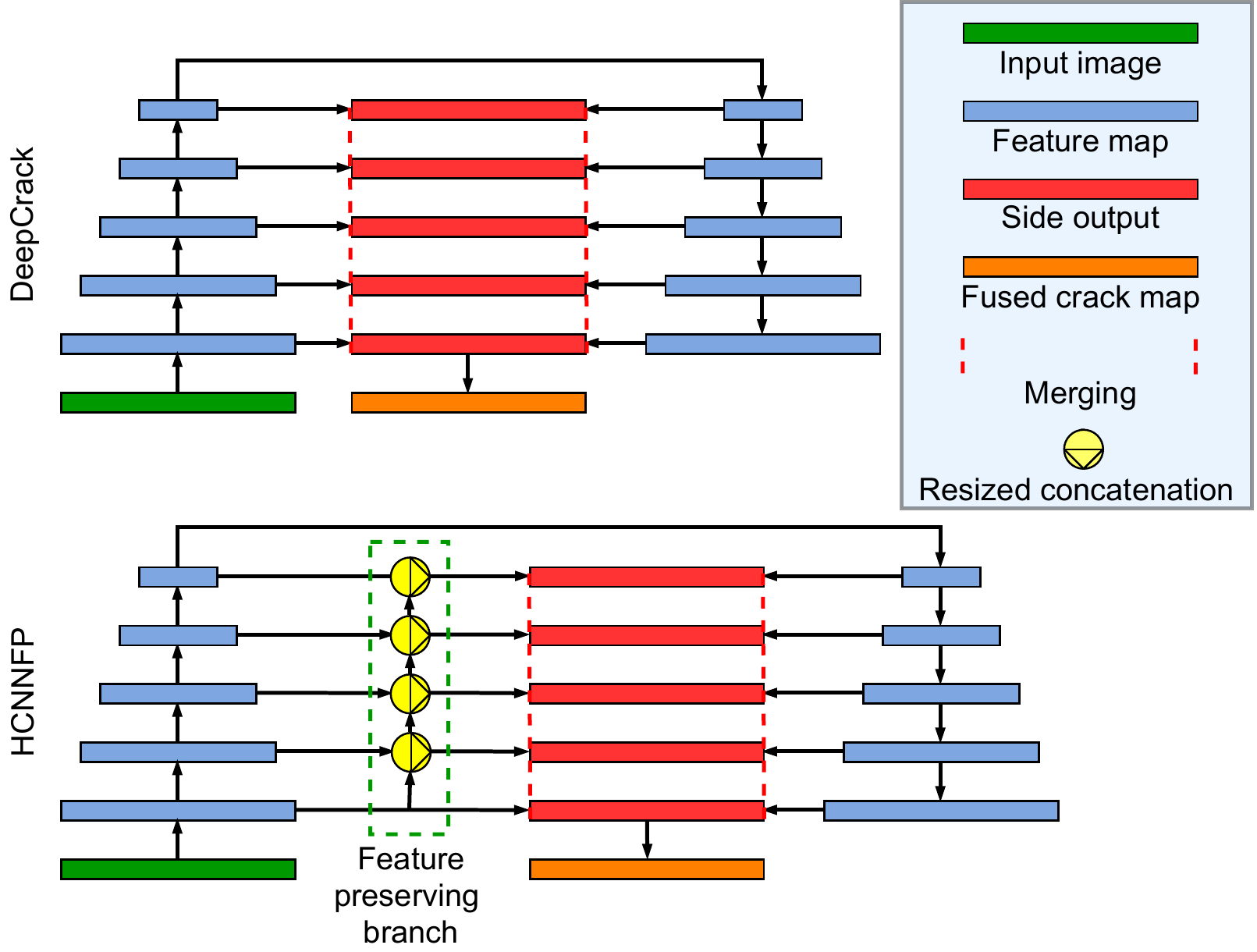}
		\caption{Architecture comparison}
		\label{fig:compare}
	\end{figure}

	By considering the benefits in the alleviation of redundant nonlinearity, here a side branch is created in the original HCNN to adjust the abstraction weights from the upper-layer by concatenation with the downsampling layer.
	 As can be seen from the proposed network architecture of Fig. \ref{fig:net}, a part of the encoder output passes by convolution, batch normalization and ReLU through an extra path and is half-sizedly concatenated with max-pooling in the downsampling. Furthermore, the concatenation takes place recursively between the output from the shallower encoder block and the output at the next deeper block. The input from each encoder block keeps semi-inherited in the feature merging routine to increase the possession of shallower-level features in merging channels along with the propagation in the convolutional network. 
	A comparison of Deepcrack architecture and our proposed one is shown in Fig. \ref{fig:compare}, wherein learning performance can be significantly improved from resized concatenations in FPB so as not to increase the computational latency. \\
	
	Notably, in our proposed enhanced HCNN network, the encoder outputs place higher weights on the feature maps from shallow layers, which result in (i) feature preservation via reduced nonlinearity as per Eq. (12), (ii) maintenance of the performance of information processing via deep learning, and (iii) assurance of the same level of computational complexity via half-sized concatenation. Under the premise of overall nonlinearity reduction and merits of the proposed approach, this adjustment can enhance the quality of the probability maps, and as such, increase the accuracy of crack detection for automatic monitoring of built infrastructure for transportation.

	\section{CONTRAST-BASED AUTOTUNED THRESHOLDING}\label{postprocessing}
	The output of the proposed pipeline is a probability map represented by a sigmoidal function. To obtain the binary map of images captured for surface inspection, it is required a suitable threshold for a fitting match to the image ground truth. The threshold, used to categorize the probability map into two classes, should be adaptable to various scenarios of surface imagery for accurate detection of cracks. To this end, an autotuning iterative thresholding technique is proposed to obtain the best threshold based on image contrast and subject to various thresholding criteria.
	
	\subsection{Probability map thresholding criteria}
	In this paper, the following criteria are used for evaluation.

	\textit{F-measure ($F_\beta$)}:
	The probability map is binarized by a threshold $T \in(0,1)$, where pixels with a probability greater (smaller) than $T$ are categorized as imperfect or intact regions. For a single image, the score $F_\beta$ \cite{borji2015salient} is often used as a fitness criterion to justify the chosen threshold. For a well-selected threshold, such F-measure is expected to reach the maximum. It is calculated as 
	\begin{equation}
	F_\beta = (1+\beta^2)\cdot \frac{ { p_T} \times  {r_T}}{ {\beta^2\times p_T}+ {r_T}},
	\label{eq:f}
	\end{equation}
	where $p_T$ and $r_T$ are respectively the precision and recall, based on the correctly-reported and falsely-reported positive or negative results; and $\beta$ denotes the weight between precision $p_T$ and recall $r_T$. A larger F-measure indicates a higher performance of the thresholding. In industrial practice, precision plays an important role in further disposal since $p_T$ score represents the ratio between the identified defect over and the return features. Such information is quite valuable to the decision on the scope of maintenance or repair work to remedy the identified defect. 
	
	\textit{Average F-measure ($AF_\beta$)}:
	To emphasize the precision over recall $r_T$, the weight $\beta^2$ should be chosen less than 1. Especially, when $\beta^2=1$, $F_\beta$ becomes the standard F-measure with equal weighting on the precision and recall \cite{borji2015salient}, which can be expressed as $F_1=2\times IoU/(1+IoU)$ and considered as mathematically alternative to the Intersection-over-Union (IoU) metric itself \cite{rezatofighi2019generalized}. As recommended in \cite{cheng2014global}, coefficient $\beta^2$ could be selected at 0.3. On other hand, $\beta^2$ = 0.25 is also frequently used to evaluate the quality of image processing \cite{milan2018semantic}. However, no strong evidence is demonstrated in the literature to prove the priority of 0.25 or 0.3 among other adjacent values. For a fairer comparison through the F-measure, we propose a new evaluation metric to calculate the average F-measure $AF_\beta$ over a given range of $\beta^2$. Here, the average F-measure, $AF_\beta$, is formulated by:
	\begin{equation}
	AF_\beta = \frac{1}{\beta^2_2-\beta^2_1}\int_{\beta^2_1}^{\beta^2_2} F_\beta d\beta^2,
	\label{eq:AF}
	\end{equation}
	where $\beta_1^2$ and $\beta_2^2$ represent are respectively the lower and upper limit for the interested range of weight $\beta^2$.
	Substituting Eq. (\ref{eq:f}) to Eq. (\ref{eq:AF}), $AF_\beta$ can be explicitly obtained as
	\begin{equation}
	AF_\beta = r_T + \frac{r_T}{p_T}\times \frac{p_T-r_T}{\beta^2_2-\beta^2_1} \times \ln \frac{p_T\beta^2_2+r_T}{p_T\beta^2_1+r_T}.
	\label{eq:baf} 
	\end{equation}
	
	Notably, this metric is considered as robust over any range of interest for weight $\beta^2$. Here, $\beta^2$ is nonzero while the condition that $\beta^2<1$ should be kept as in common practice. In terms of evaluation judgment, similarly to $F_\beta$, a higher $AF_\beta$ represents better quality of a thresholding technique.  
	
	\textit{MAE}: 
	For a binary map $S=\{s_{ij}\}$, the mean absolute error (MAE) can be obtained from the post-processing step as, 
	\begin{equation}
	MAE=\frac{1}{I \times J} \sum_{i=1}^{I}\sum_{j=1}^{J}|s_{ij}-y_{ij}|.
	\end{equation}  
	A smaller \textit{MAE} indicates a better match to the ground truth (GT). Complementarily to the weighted F-measure, this metric rewards predictions with a high recall rate due to its preference for false-positive samples \cite{hou2017deeply}.  Alternatively, one can also use the mean absolute percentage error (MAPE) \cite{de2016mean} with the total number of pixels in the image being replaced by the number of true-positive pixels $N_{tp}$ in the denominator of MAE as,
	\begin{equation}
	MAPE=\frac{1}{N_{tp}} \sum_{i=1}^{I}\sum_{j=1}^{J}|s_{ij}-y_{ij}|
	\end{equation}
	to make the results more salient.
	
	\subsection{Intercontrast iterative thresholding}
\begin{figure}[h!]
		\centering
		\includegraphics[width=0.35\textwidth]{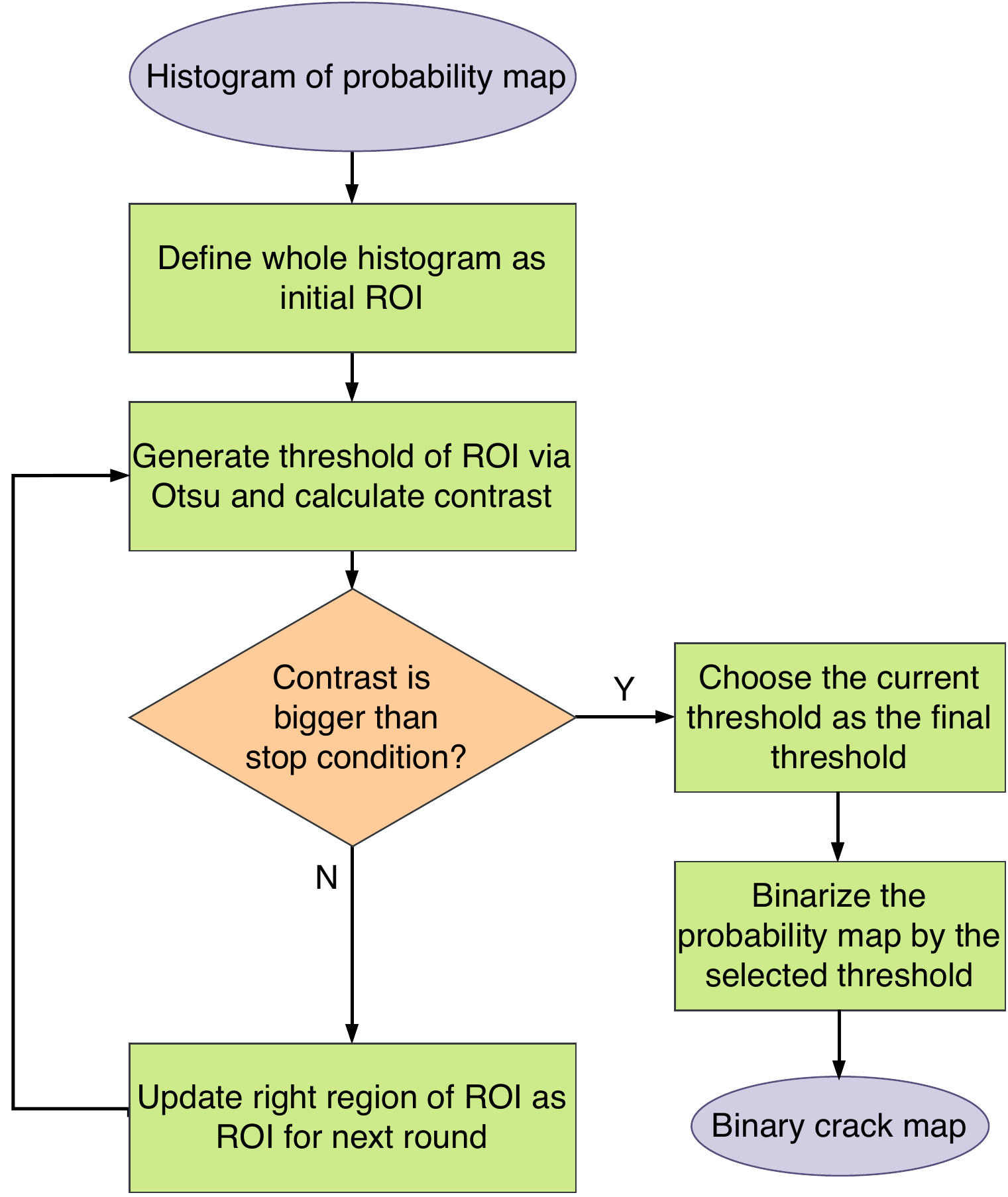}
		\caption{Binarization flowchart using CBAT}
		\label{fig:flowchart}
	\end{figure}

	An even threshold $T=0.5$ is usually considered as a reasonable value for good thresholding. However, this may cause mislabeling in the case with an unbalanced ratio between a faulty feature and an intact background. To obtain a better result, it is worth seeking a mechanism for autotuning of the threshold. To this end, we propose the contrast-based autotuned thresholding (CBAT), a contrast-based approach refined from Otsu's thresholding \cite{otsu1979threshold}, to improve the accuracy of binarization. A flowchart for obtaining the binary map is depicted in Fig. \ref{fig:flowchart}, wherein Otsu's thresholding is only implemented in the region of interest (ROI) that encompasses a cluster of high-probability defect candidates during the iteration process rather than a large background region.  

	In the initial iteration, the whole histogram is predefined as the original ROI $R^0_{ROI}$.
	Generally, in the $m^{th}$ iteration, Otsu's algorithm $otsu(.)$ obtains a threshold $T^m_{ROI}$ for the previous ROI $R^{m-1}_{ROI}$ such that
	\begin{equation}
	otsu(R^{m-1}_{ROI})=  T^m_{ROI}. 
	\end{equation}  
	Threshold $T^m_{ROI}$ is expected to segment $R^{m-1}_{ROI}$ into region of interest $R^m_{ROI}$ and background $R^m_b$, i.e.
	\begin{equation}
	R^{m-1}_{ROI} =  R^m_{ROI} \cup R^m_b, 
	\end{equation}
	where $R^m_{ROI}$ is the current ROI containing the pixels with a probability higher than $T^m_{ROI}$, and $R^m_b$ is the corresponding background region for pixels whose probability is between $T^m_{ROI}$ and $T^{m-1}_{ROI}$. 
	
	The interclass contrast \cite{levine:measurement} is a criterion for the assessment of segmentation quality, under the hypothesis that the intensity of homogenous pixels is close to the average intensity of their class. Referred to the probability histogram, the interclass contrast $C^m_{ROI}$ for region $R^{m-1}_{ROI}$ is expressed as,
	\begin{equation}
	C^m_{ROI} =\frac{\mid\mu^m_{ROI}-\mu^m_b\mid}{\mu^m_{ROI}+\mu^m_b},
	\end{equation}
	where $\mu^m_{ROI}$ and $\mu^m_b$ are respectively the mean probability in $R^m_{ROI}$ and $R^m_b$. Due to a significant reduction of the ROI pixel number, the sum $\mu^m_{ROI}+\mu^m_b$ remains decreasing with iterations.
	Consequently, $C^m_{ROI}$ keeps increasing until the iteration terminates. A strong contrast implies an obvious difference between the two classes within the probability map, resulting in a distinction between abnormal features and the intact region on the image captured.
	 
	Since our target is to highlight an imperfect region as a crack from its neighborhood, a proper interclass contrast is required to preserve the discernibility of defect candidates. For a particular surface type, the termination condition for contrast $C_s$ is set so that the loop will stop when $C^m_{ROI}>C_s$ to yield the ultimate threshold $T_u$. 
	The pseudo-code for the proposed CBAT approach is demonstrated in Algorithm \ref{alg:frontthresh}.

	\begin{algorithm}
		\caption{Intercontrast Iterative Thresholding}
		\label{alg:frontthresh}
		\begin{algorithmic}[1]
			\Require  $R^0_i$: whole histogram
			\Ensure $T_u$: ultimate threshold
			\State  $m \gets 0$
			\Repeat
			\State   $m++$
			\State  $T^m_{ROI} \gets otsu(R^{m-1}_{ROI})$
			\State  $R^m_{ROI} \gets R^{m-1}_i(R^{m-1}_{ROI}>T^m_{ROI})$
			\State $R^m_b \gets R^{m-1}_{ROI}(T^{m-1}_{ROI}\leq R^{m-1}_{ROI}<T^m_{ROI})$
			\State $\mu^m_{ROI} \gets Average(R^m_{ROI})$, $\mu^m_b \gets Average(R^m_b)$
			\State $C^m_{ROI} \gets {\mid\mu^m_{ROI}-\mu^m_b\mid}/(\mu^m_{ROI}+\mu^m_b)$
			\Until {$(C^m_{ROI}>C_s)$}
			\State $T_u \gets T^m_{ROI}$			
		\end{algorithmic}
	\end{algorithm}
	
The ROI, initially the whole histogram, is reduced after running the Otsu's algorithm for the first time at threshold $T_1$, then keeps shrinking iteratively with an updated interclass contrast $(C^m_{ROI}$ to result in a new region of interest during the next searches for $T_2,~T_3, ...$ until reaching the termination condition for the ultimate value $T_u$. The probability region with a threshold $T$ greater than $T_u$ will be assigned to defect candidates, and the remaining pixels will be categorized as the background. 
	
	\section{Experiments and Evaluation}\label{experiment}
\begin{figure}[h]
	\centering	
	\includegraphics[width=0.45\textwidth]{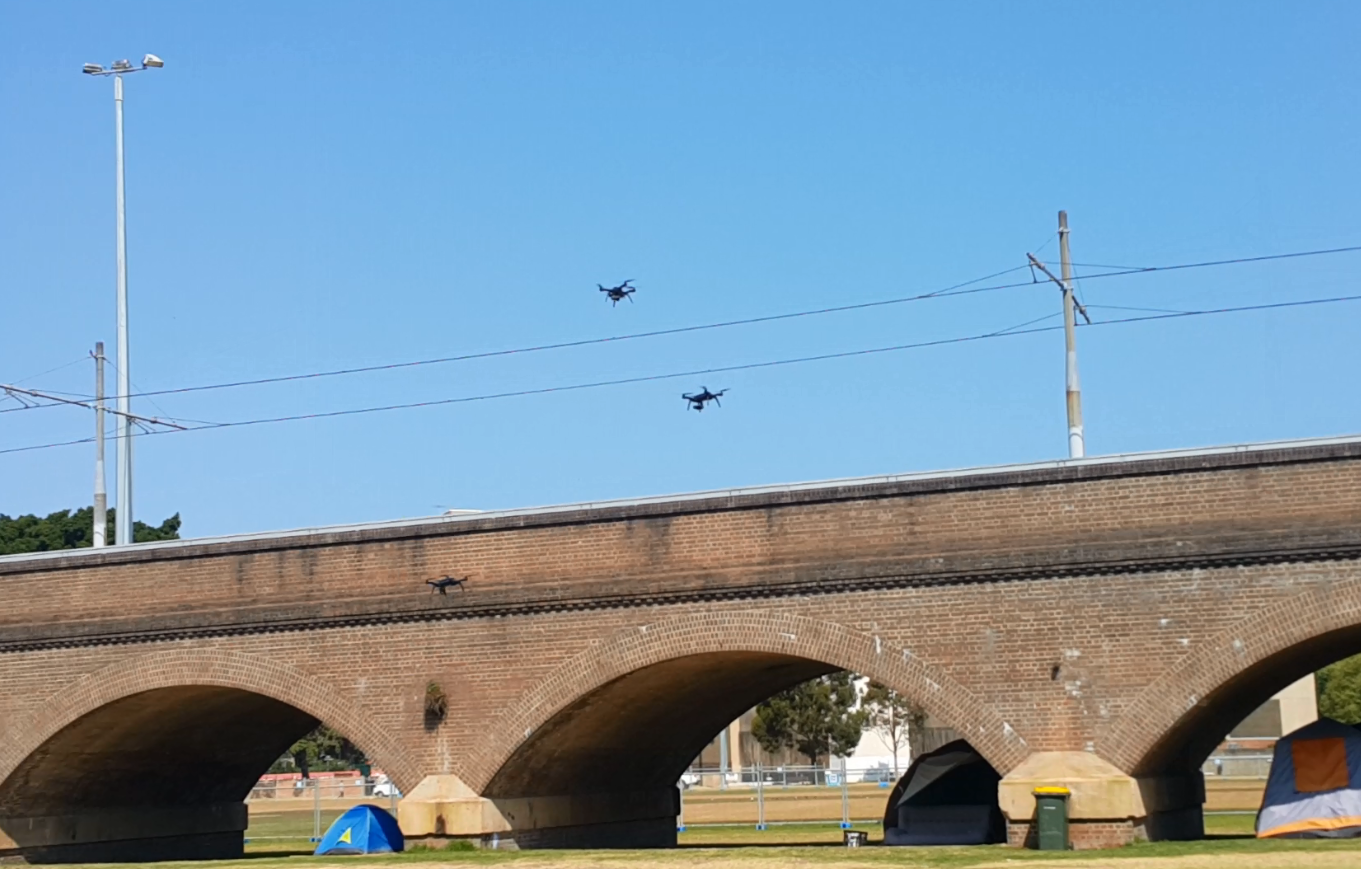}
	\caption{Bridge inspection}
	\label{fig:formation}
\end{figure}

The UAV inspection system used for collecting images 
includes two subsystems: Skynet for flying drones and capturing images, and the ground center for data processing. The quadcopters are controlled to follow an IoT-communicated formation \cite{hoang2019reconfigurable} to inspect a monorail bridge, as depicted in Fig. \ref{fig:formation}. Here, all the trainable parameters are initialized by He Normal initialization \cite{he2015delving}. The training is conducted at a learning rate of 1$e$-5 and optimized by Adaptive Moment Estimation (Adam) \cite{kingma2014adam} with the default setting of two hyperparameters, (0.9, 0.999). For performance improvement from using the proposed loss function, the maximum training epoch is set as 30, which is adequately large for the sake of convergence. An early stop is applied when the reduction of the loss between two adjacent epochs is under 0.01\%. The training process is conducted on NVIDIA Tesla T4 GPUs 16Gb.    

\subsubsection{Datasets}
Four datasets used in our experiments include:
\begin{itemize}
	\item \textit{Crack500} \cite{yang2019feature}: containing 500 images of pavement cracks with granular backgrounds in a unified size of $2560 \times 1440$ with a few samples under uneven illumination. Due to the limitation in GPU memory and computation power, all the images are resized to $256 \times 256$. 
	
	\item \textit{CrackForest}\cite{shi2016automatic}: containing 118 images of road cracks with labeled masks in a size of $600 \times 800$ with a part of samples with the interference of shadow and painted marks. We rotate the images with a range from 0 to 90 degrees, flip them vertically and horizontally, and randomly crop the flipped images with a size of $256 \times 256$. Ten thousands augmented images are split into the training and the validation set with a ratio of 9:1. The rest 1180 images are preserved for testing.

	\item \textit{DCD} \cite{liu2019deepcrack}: containing 521 images of infrastructure cracks with texture and misleading marks under various light condition. 
	
	\item \textit{GAPs} \cite{2017get}: containing 509 images of pavement cracks with densely granular backgrounds under poor light conditions. DCD and GAPs are both integrated into an unified size of $448 \times 448$, following  \cite{liu2019deep}. 
\end{itemize}
Here, the original annotation of public datasets is kept for a fair comparison with peer methods. For images with larger sizes, a sliding window can be used to process the image clip by clip \cite{pan2020new}.  Our original dataset is also included for testing:
\begin{itemize}
	\item \textit{SYDCrack} \cite{Zhu2019}: With the inspection system introduced above, an image dataset is collected for some surface cracks on a monorail bridge with regularly textured backgrounds under a good illumination. Those images are collected at 15 locations where crack patterns are located. The integrated dataset contains 170 images, cropped into 850 pitches with a size of $224 \times 224$.
	
\end{itemize}

	\subsubsection{Benchmarking}\label{method}

In the first experiment, a comparative analysis is conducted between our proposed one and the recent state-of-the-art crack detection methods using deep learning. The frameworks for comparison are listed in the following:	
	\begin{itemize} 
		\item \textit{HED}\cite{xie2015holistically}:
		Retaining the encoder part of SegNet, the holistically nested edge detection (HED) merges feature maps from five different levels in the encoder. The last feature map is used for computation of the loss function.   
		
		\item \textit{RCF}\cite{liu2017richer}:
		Another edge detection technique for richer convolutional features (RCF) delivers a merged output from each convolutional layer in the encoder block while HED just outputs the final layer.
		
		\item \textit{SegNet}\cite{badrinarayanan2017segnet}: This framework represents a standard end-to-end model with an auto-encoder.  
		
		\item \textit{DeepCrack}\cite{zou2018deepcrack}: An end-to-end hierarchical network for crack extraction using the typical architecture of SegNet with symmetrical concatenation in the side branch. 
		\item \textit{FPHBN}\cite{yang2019feature}: The feature pyramid and hierarchical boosting network (FPHBN) is a recently proposed framework for crack detection and constructed on the main structure of HED.
		
		\item \textit{FCN}\cite{dung2019autonomous}:
		A simplified hourglass shape network for crack detection with only 6 blocks.   
		
		\item \textit{U-Net}\cite{liu2019computer}: An U-shape auto-encoder network with scale-invariant merging between outputs from the encoder and the decoder.

	\end{itemize}

\begin{table*}[tbh]
	\renewcommand{\arraystretch}{1.3}
	\footnotesize\addtolength{\tabcolsep}{-5pt}
	\begin{center}
		\begin{tabular}{cllllllll}	
			Crack500-1\hspace{0.15cm}  & \begin{subfigure}{0.086\textwidth}\centering\includegraphics[width=\linewidth]{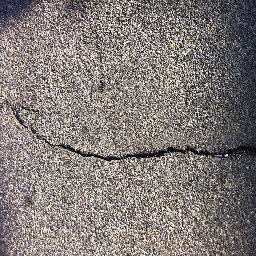}\label{fig:taba}\end{subfigure}
			\begin{subfigure}{0.086\textwidth}\centering\includegraphics[width=\linewidth]{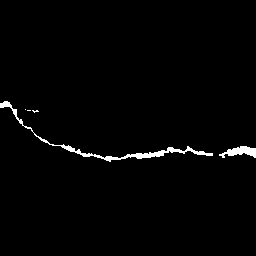}\label{fig:taba}\end{subfigure}
			\begin{subfigure}{0.086\textwidth}\centering\includegraphics[width=\linewidth]{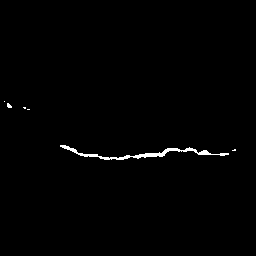}\label{fig:tabb}\end{subfigure}
			\begin{subfigure}{0.086\textwidth}\centering\includegraphics[width=\linewidth]{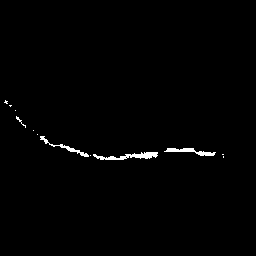}\label{fig:taba}\end{subfigure}
			\begin{subfigure}{0.086\textwidth}\centering\includegraphics[width=\linewidth]{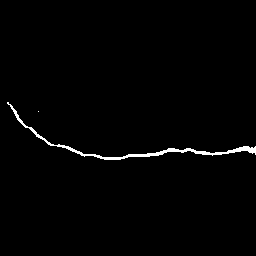}\label{fig:tabc}\end{subfigure}
			\begin{subfigure}{0.086\textwidth}\centering\includegraphics[width=\linewidth]{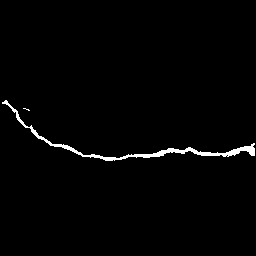}\label{fig:tabc}\end{subfigure}
			\begin{subfigure}{0.086\textwidth}\centering\includegraphics[width=\linewidth]{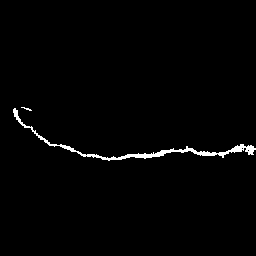}\label{fig:tabc}\end{subfigure}
			\begin{subfigure}{0.086\textwidth}\centering\includegraphics[width=\linewidth]{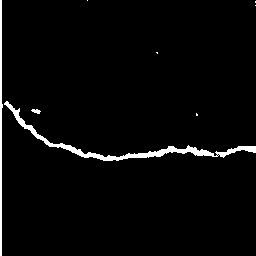}\label{fig:tabc}\end{subfigure}
			\begin{subfigure}{0.086\textwidth}\centering\includegraphics[width=\linewidth]{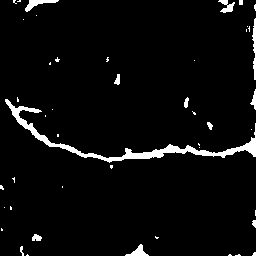}\label{fig:tabc}\end{subfigure}
			\begin{subfigure}{0.086\textwidth}\centering\includegraphics[width=\linewidth]{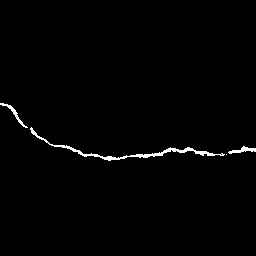}\label{fig:tabc}\end{subfigure}\vspace{3px}\\[3mm]
			Crack500-2 \hspace{0.15cm} & \begin{subfigure}{0.086\textwidth}\centering\includegraphics[width=\linewidth]{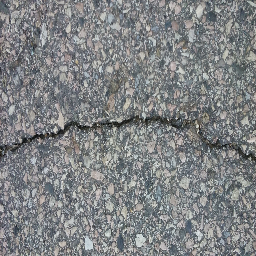}\label{fig:taba}\end{subfigure}
			\begin{subfigure}{0.086\textwidth}\centering\includegraphics[width=\linewidth]{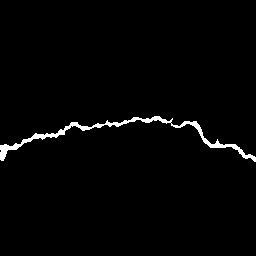}\label{fig:taba}\end{subfigure}
			\begin{subfigure}{0.086\textwidth}\centering\includegraphics[width=\linewidth]{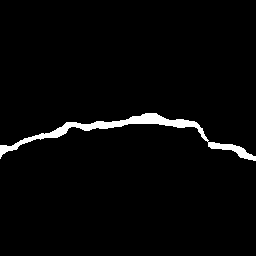}\label{fig:tabb}\end{subfigure}
			\begin{subfigure}{0.086\textwidth}\centering\includegraphics[width=\linewidth]{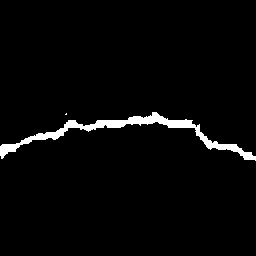}\label{fig:taba}\end{subfigure}
			\begin{subfigure}{0.086\textwidth}\centering\includegraphics[width=\linewidth]{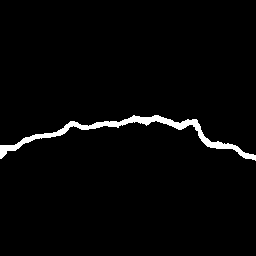}\label{fig:tabc}\end{subfigure}
			\begin{subfigure}{0.086\textwidth}\centering\includegraphics[width=\linewidth]{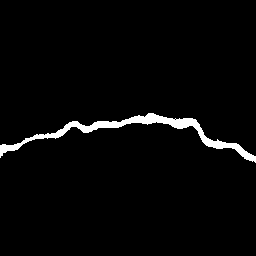}\label{fig:tabc}\end{subfigure}
			\begin{subfigure}{0.086\textwidth}\centering\includegraphics[width=\linewidth]{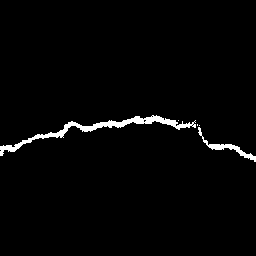}\label{fig:tabc}\end{subfigure}
			\begin{subfigure}{0.086\textwidth}\centering\includegraphics[width=\linewidth]{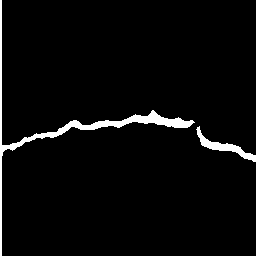}\label{fig:tabc}\end{subfigure}
			\begin{subfigure}{0.086\textwidth}\centering\includegraphics[width=\linewidth]{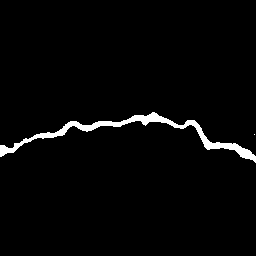}\label{fig:tabc}\end{subfigure}
			\begin{subfigure}{0.086\textwidth}\centering\includegraphics[width=\linewidth]{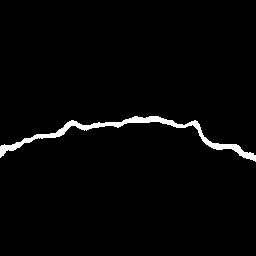}\label{fig:tabc}\end{subfigure}\vspace{3px}\\[3mm]
			CrackForest-3 \hspace{0.05cm} &		
			\begin{subfigure}{0.086\textwidth}\centering\includegraphics[width=\linewidth]{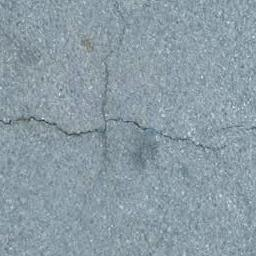}\label{fig:taba}\end{subfigure}
			\begin{subfigure}{0.086\textwidth}\centering\includegraphics[width=\linewidth]{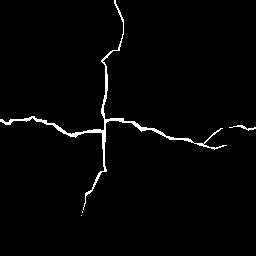}\label{fig:taba}\end{subfigure}
			\begin{subfigure}{0.086\textwidth}\centering\includegraphics[width=\linewidth]{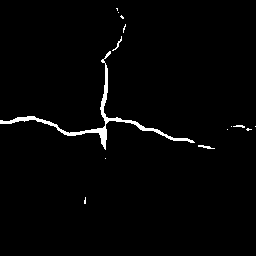}\label{fig:tabb}\end{subfigure}
			\begin{subfigure}{0.086\textwidth}\centering\includegraphics[width=\linewidth]{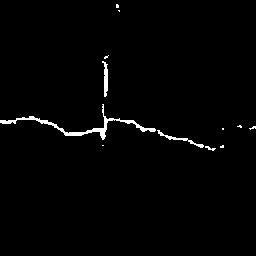}\label{fig:taba}\end{subfigure}
			\begin{subfigure}{0.086\textwidth}\centering\includegraphics[width=\linewidth]{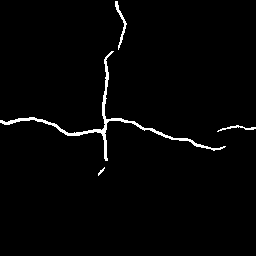}\label{fig:tabc}\end{subfigure}
			\begin{subfigure}{0.086\textwidth}\centering\includegraphics[width=\linewidth]{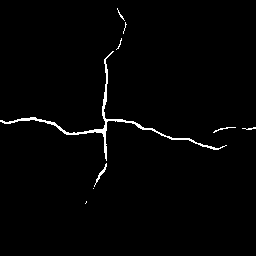}\label{fig:tabc}\end{subfigure}
			\begin{subfigure}{0.086\textwidth}\centering\includegraphics[width=\linewidth]{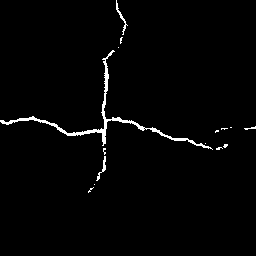}\label{fig:tabc}\end{subfigure}
			\begin{subfigure}{0.086\textwidth}\centering\includegraphics[width=\linewidth]{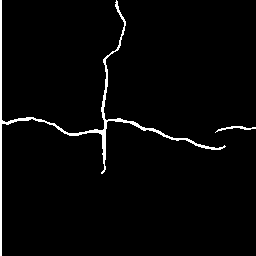}\label{fig:tabc}\end{subfigure}
			\begin{subfigure}{0.086\textwidth}\centering\includegraphics[width=\linewidth]{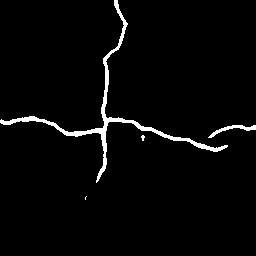}\label{fig:tabc}\end{subfigure}
			\begin{subfigure}{0.086\textwidth}\centering\includegraphics[width=\linewidth]{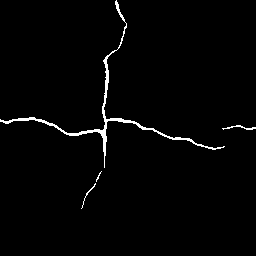}\label{fig:tabc}\end{subfigure}\vspace{3px}\\[3mm]
			CrackForest-4 \hspace{0.05cm} &	\begin{subfigure}{0.086\textwidth}\centering\includegraphics[width=\linewidth]{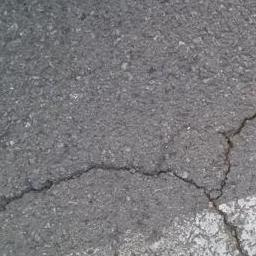}\label{fig:taba}\end{subfigure}
			\begin{subfigure}{0.086\textwidth}\centering\includegraphics[width=\linewidth]{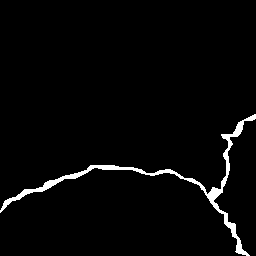}\label{fig:taba}\end{subfigure}
			\begin{subfigure}{0.086\textwidth}\centering\includegraphics[width=\linewidth]{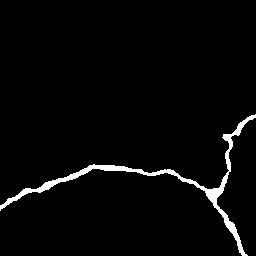}\label{fig:tabb}\end{subfigure}
			\begin{subfigure}{0.086\textwidth}\centering\includegraphics[width=\linewidth]{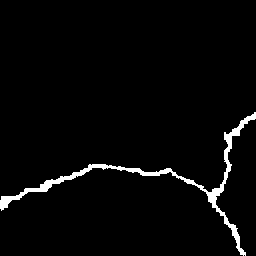}\label{fig:taba}\end{subfigure}
			\begin{subfigure}{0.086\textwidth}\centering\includegraphics[width=\linewidth]{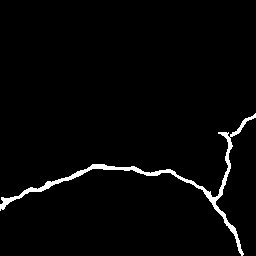}\label{fig:tabc}\end{subfigure}
			\begin{subfigure}{0.086\textwidth}\centering\includegraphics[width=\linewidth]{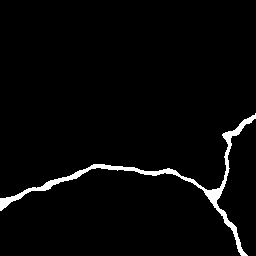}\label{fig:tabc}\end{subfigure}
			\begin{subfigure}{0.086\textwidth}\centering\includegraphics[width=\linewidth]{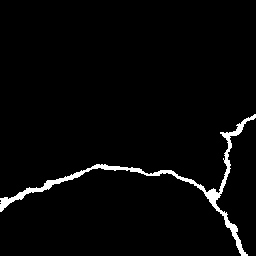}\label{fig:tabc}\end{subfigure}
			\begin{subfigure}{0.086\textwidth}\centering\includegraphics[width=\linewidth]{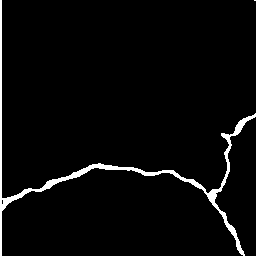}\label{fig:tabc}\end{subfigure}
			\begin{subfigure}{0.086\textwidth}\centering\includegraphics[width=\linewidth]{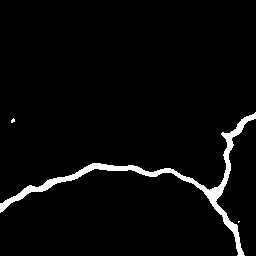}\label{fig:tabc}\end{subfigure}
			\begin{subfigure}{0.086\textwidth}\centering\includegraphics[width=\linewidth]{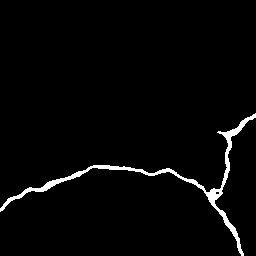}\label{fig:tabc}\end{subfigure}\vspace{3px}\\[3mm]	
			SYDCrack-5 \hspace{0.15cm} & \begin{subfigure}{0.086\textwidth}\centering\includegraphics[width=\linewidth]{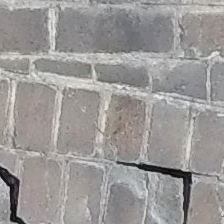}\label{fig:taba}\end{subfigure}
			\begin{subfigure}{0.086\textwidth}\centering\includegraphics[width=\linewidth]{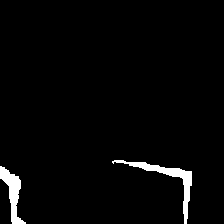}\label{fig:taba}\end{subfigure}
			\begin{subfigure}{0.086\textwidth}\centering\includegraphics[width=\linewidth]{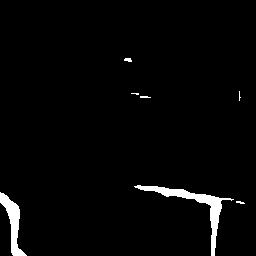}\label{fig:tabb}\end{subfigure}
			\begin{subfigure}{0.086\textwidth}\centering\includegraphics[width=\linewidth]{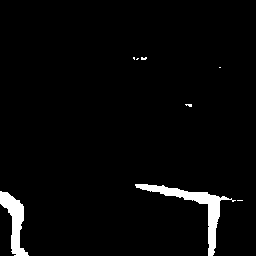}\label{fig:taba}\end{subfigure}
			\begin{subfigure}{0.086\textwidth}\centering\includegraphics[width=\linewidth]{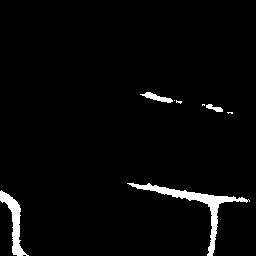}\label{fig:tabc}\end{subfigure}
			\begin{subfigure}{0.086\textwidth}\centering\includegraphics[width=\linewidth]{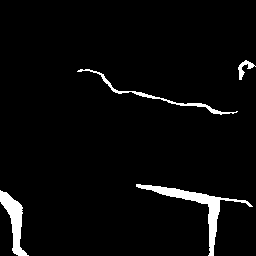}\label{fig:tabc}\end{subfigure}
			\begin{subfigure}{0.086\textwidth}\centering\includegraphics[width=\linewidth]{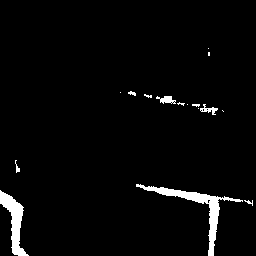}\label{fig:tabc}\end{subfigure}
			\begin{subfigure}{0.086\textwidth}\centering\includegraphics[width=\linewidth]{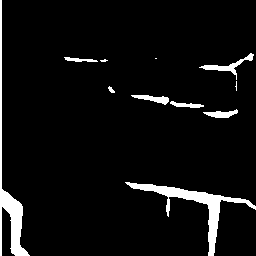}\label{fig:tabc}\end{subfigure}
			\begin{subfigure}{0.086\textwidth}\centering\includegraphics[width=\linewidth]{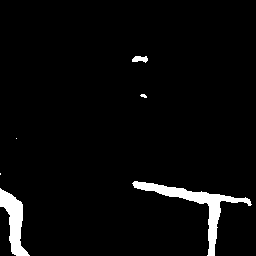}\label{fig:tabc}\end{subfigure}		
			\begin{subfigure}{0.086\textwidth}\centering\includegraphics[width=\linewidth]{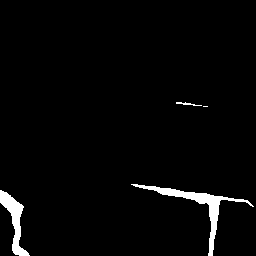}\label{fig:tabc}\end{subfigure}\vspace{3px}\\[3mm]
			
			SYDCrack-6\hspace{0.15cm} &
			\begin{subfigure}{0.086\textwidth}\centering\includegraphics[width=\linewidth]{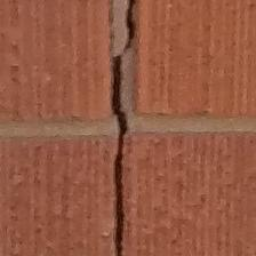}\label{fig:taba}\end{subfigure}		
			\begin{subfigure}{0.086\textwidth}\centering\includegraphics[width=\linewidth]{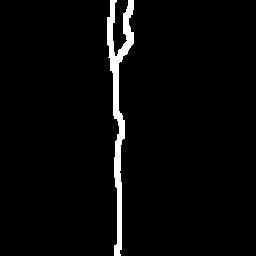}\label{fig:taba}\end{subfigure}
			\begin{subfigure}{0.086\textwidth}\centering\includegraphics[width=\linewidth]{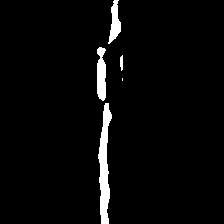}\label{fig:taba}\end{subfigure}
			\begin{subfigure}{0.086\textwidth}\centering\includegraphics[width=\linewidth]{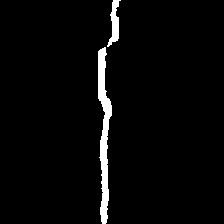}\label{fig:taba}\end{subfigure}
			\begin{subfigure}{0.086\textwidth}\centering\includegraphics[width=\linewidth]{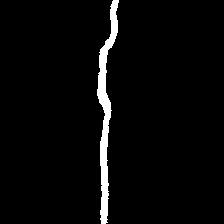}\label{fig:tabc}\end{subfigure}
			\begin{subfigure}{0.086\textwidth}\centering\includegraphics[width=\linewidth]{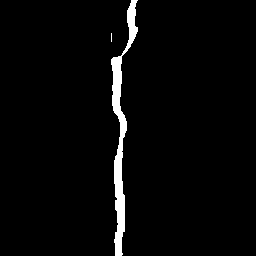}\label{fig:taba}\end{subfigure}
			\begin{subfigure}{0.086\textwidth}\centering\includegraphics[width=\linewidth]{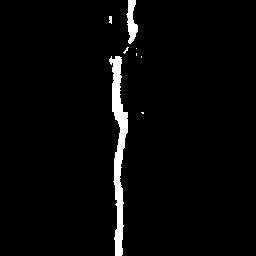}\label{fig:tabc}\end{subfigure}
			\begin{subfigure}{0.086\textwidth}\centering\includegraphics[width=\linewidth]{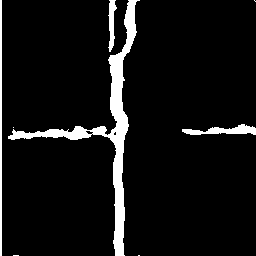}\label{fig:tabb}\end{subfigure}
			\begin{subfigure}{0.086\textwidth}\centering\includegraphics[width=\linewidth]{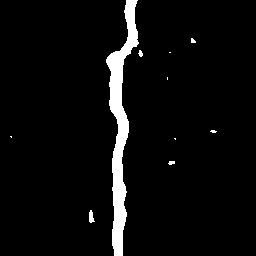}\label{fig:taba}\end{subfigure}
			\begin{subfigure}{0.086\textwidth}\centering\includegraphics[width=\linewidth]{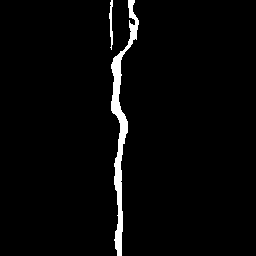}\label{fig:tabc}\end{subfigure}\vspace{3px}\\[3mm]
			
			DCD-7\hspace{0.15cm}&
			\begin{subfigure}{0.086\textwidth}\centering\includegraphics[width=\linewidth]{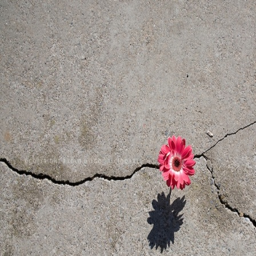}\label{fig:taba}\end{subfigure}		
			\begin{subfigure}{0.086\textwidth}\centering\includegraphics[width=\linewidth]{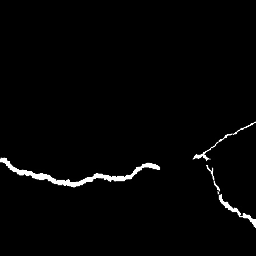}\label{fig:taba}\end{subfigure}
			\begin{subfigure}{0.086\textwidth}\centering\includegraphics[width=\linewidth]{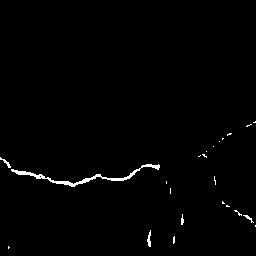}\label{fig:taba}\end{subfigure}
			\begin{subfigure}{0.086\textwidth}\centering\includegraphics[width=\linewidth]{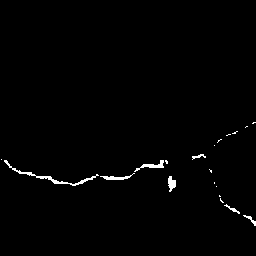}\label{fig:taba}\end{subfigure}
			\begin{subfigure}{0.086\textwidth}\centering\includegraphics[width=\linewidth]{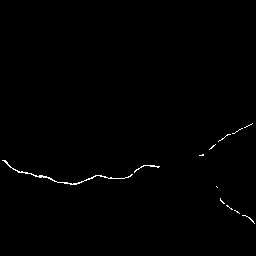}\label{fig:tabc}\end{subfigure}
			\begin{subfigure}{0.086\textwidth}\centering\includegraphics[width=\linewidth]{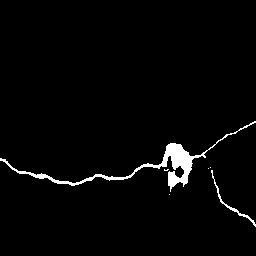}\label{fig:taba}\end{subfigure}
			\begin{subfigure}{0.086\textwidth}\centering\includegraphics[width=\linewidth]{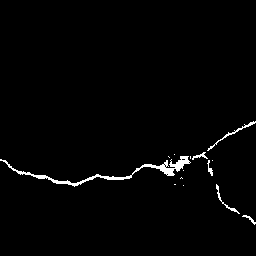}\label{fig:tabc}\end{subfigure}
			\begin{subfigure}{0.086\textwidth}\centering\includegraphics[width=\linewidth]{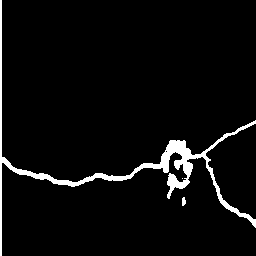}\label{fig:tabb}\end{subfigure}
			\begin{subfigure}{0.086\textwidth}\centering\includegraphics[width=\linewidth]{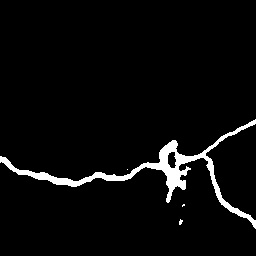}\label{fig:taba}\end{subfigure}
			\begin{subfigure}{0.086\textwidth}\centering\includegraphics[width=\linewidth]{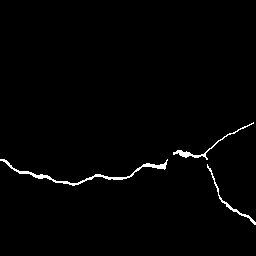}\label{fig:tabc}\end{subfigure}\vspace{3px}\\[3mm]
			
			DCD-8\hspace{0.15cm}&
			\begin{subfigure}{0.086\textwidth}\centering\includegraphics[width=\linewidth]{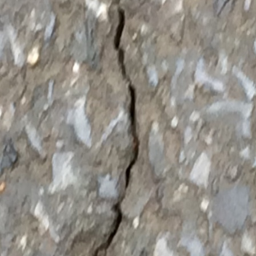}\label{fig:taba}\end{subfigure}		
			\begin{subfigure}{0.086\textwidth}\centering\includegraphics[width=\linewidth]{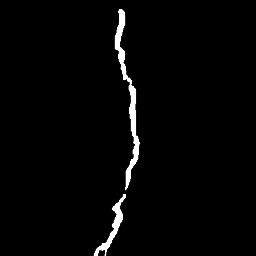}\label{fig:taba}\end{subfigure}
			\begin{subfigure}{0.086\textwidth}\centering\includegraphics[width=\linewidth]{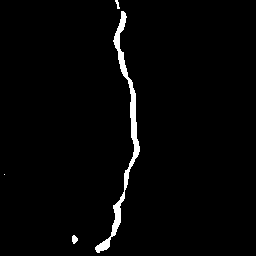}\label{fig:taba}\end{subfigure}
			\begin{subfigure}{0.086\textwidth}\centering\includegraphics[width=\linewidth]{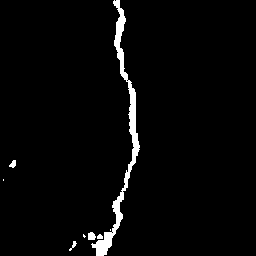}\label{fig:taba}\end{subfigure}
			\begin{subfigure}{0.086\textwidth}\centering\includegraphics[width=\linewidth]{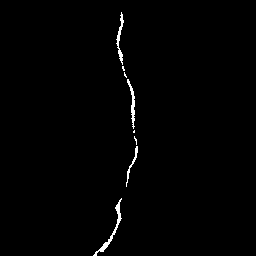}\label{fig:tabc}\end{subfigure}
			\begin{subfigure}{0.086\textwidth}\centering\includegraphics[width=\linewidth]{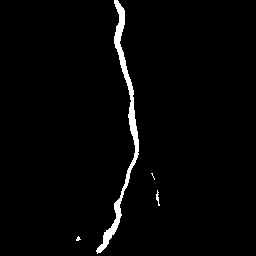}\label{fig:taba}\end{subfigure}
			\begin{subfigure}{0.086\textwidth}\centering\includegraphics[width=\linewidth]{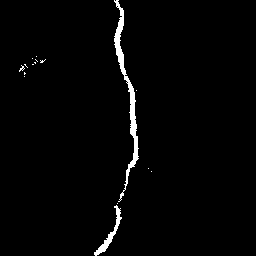}\label{fig:tabc}\end{subfigure}
			\begin{subfigure}{0.086\textwidth}\centering\includegraphics[width=\linewidth]{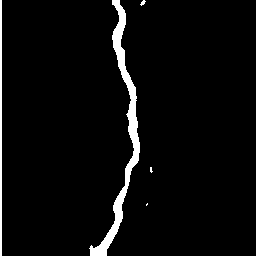}\label{fig:tabb}\end{subfigure}
			\begin{subfigure}{0.086\textwidth}\centering\includegraphics[width=\linewidth]{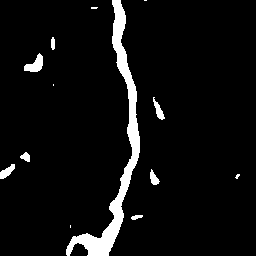}\label{fig:taba}\end{subfigure}
			\begin{subfigure}{0.086\textwidth}\centering\includegraphics[width=\linewidth]{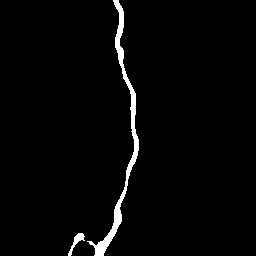}\label{fig:tabc}\end{subfigure}\vspace{3px}\\[3mm]
			
			GAPs-9\hspace{0.15cm}&
			\begin{subfigure}{0.086\textwidth}\centering\includegraphics[width=\linewidth]{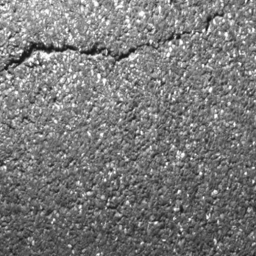}\label{fig:taba}\end{subfigure}		
			\begin{subfigure}{0.086\textwidth}\centering\includegraphics[width=\linewidth]{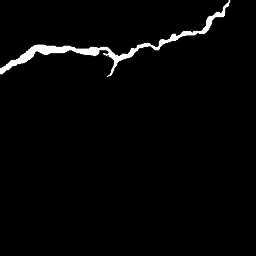}\label{fig:taba}\end{subfigure}
			\begin{subfigure}{0.086\textwidth}\centering\includegraphics[width=\linewidth]{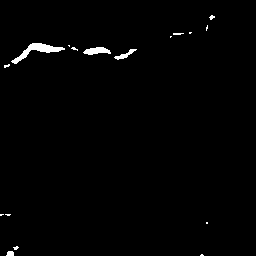}\label{fig:taba}\end{subfigure}
			\begin{subfigure}{0.086\textwidth}\centering\includegraphics[width=\linewidth]{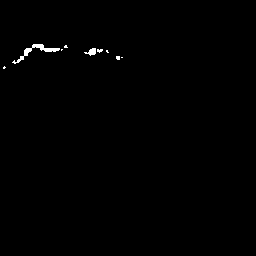}\label{fig:taba}\end{subfigure}
			\begin{subfigure}{0.086\textwidth}\centering\includegraphics[width=\linewidth]{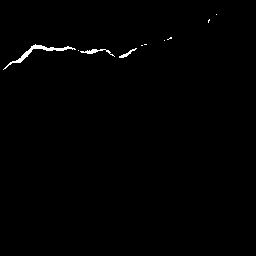}\label{fig:tabc}\end{subfigure}
			\begin{subfigure}{0.086\textwidth}\centering\includegraphics[width=\linewidth]{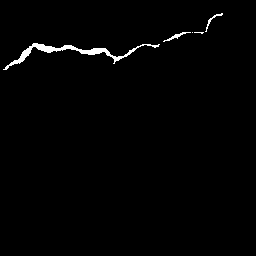}\label{fig:taba}\end{subfigure}
			\begin{subfigure}{0.086\textwidth}\centering\includegraphics[width=\linewidth]{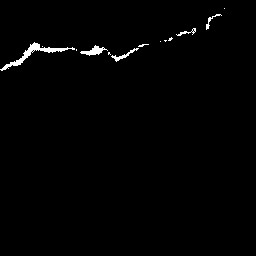}\label{fig:tabc}\end{subfigure}
			\begin{subfigure}{0.086\textwidth}\centering\includegraphics[width=\linewidth]{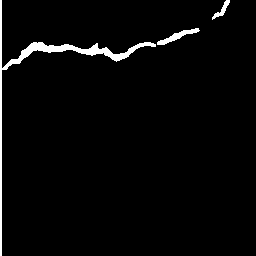}\label{fig:tabb}\end{subfigure}
			\begin{subfigure}{0.086\textwidth}\centering\includegraphics[width=\linewidth]{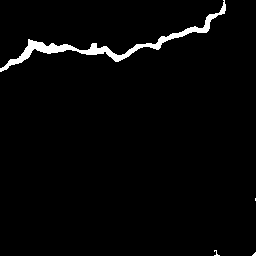}\label{fig:taba}\end{subfigure}
			\begin{subfigure}{0.086\textwidth}\centering\includegraphics[width=\linewidth]{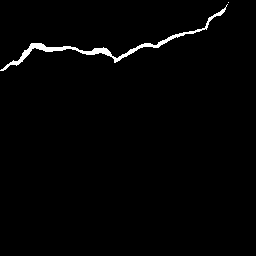}\label{fig:tabc}\end{subfigure}\vspace{3px}\\[3mm]
			
			GAPs-10\hspace{0.15cm}&
			\begin{subfigure}{0.086\textwidth}\centering\includegraphics[width=\linewidth]{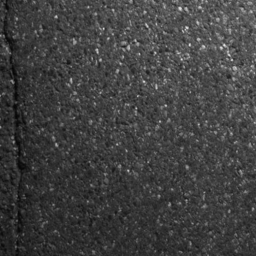}\label{fig:taba}\end{subfigure}		
			\begin{subfigure}{0.086\textwidth}\centering\includegraphics[width=\linewidth]{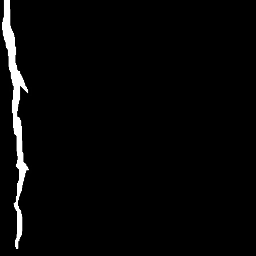}\label{fig:taba}\end{subfigure}
			\begin{subfigure}{0.086\textwidth}\centering\includegraphics[width=\linewidth]{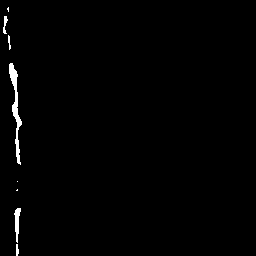}\label{fig:taba}\end{subfigure}
			\begin{subfigure}{0.086\textwidth}\centering\includegraphics[width=\linewidth]{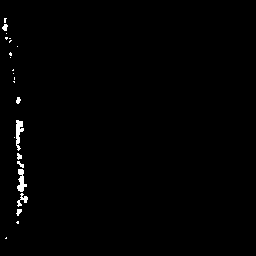}\label{fig:taba}\end{subfigure}
			\begin{subfigure}{0.086\textwidth}\centering\includegraphics[width=\linewidth]{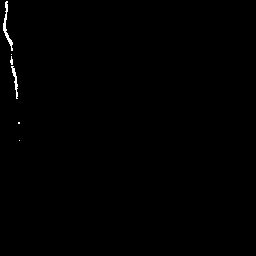}\label{fig:tabc}\end{subfigure}
			\begin{subfigure}{0.086\textwidth}\centering\includegraphics[width=\linewidth]{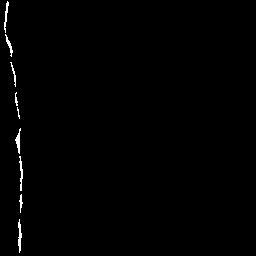}\label{fig:taba}\end{subfigure}
			\begin{subfigure}{0.086\textwidth}\centering\includegraphics[width=\linewidth]{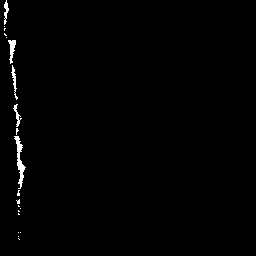}\label{fig:tabc}\end{subfigure}
			\begin{subfigure}{0.086\textwidth}\centering\includegraphics[width=\linewidth]{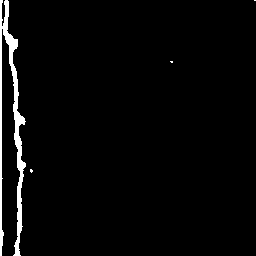}\label{fig:tabb}\end{subfigure}
			\begin{subfigure}{0.086\textwidth}\centering\includegraphics[width=\linewidth]{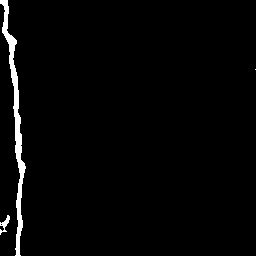}\label{fig:taba}\end{subfigure}
			\begin{subfigure}{0.086\textwidth}\centering\includegraphics[width=\linewidth]{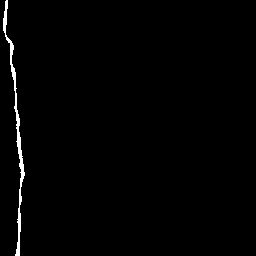}\label{fig:tabc}\end{subfigure}\vspace{3px}\\[3mm]	
			
			&\hspace{0.016\textwidth}Original\hspace{0.052\textwidth}GT\hspace{0.06\textwidth}HED\hspace{0.058\textwidth}RCF\hspace{0.054\textwidth}SegNet\hspace{0.032\textwidth}DeepCrack\hspace{0.032\textwidth}FPHBN\hspace{0.040\textwidth}U-Net\hspace{0.056\textwidth}FCN\hspace{0.042\textwidth}HCNNFP\hspace{0.038\textwidth}
			
		\end{tabular}
		\captionof{figure}{Detection results by six DCNN approaches with even threshold $T=0.5$.}\vspace{-0.35cm}
		\label{fig:comparison1}
	\end{center}
\end{table*}

	In the second experiment, the methods to compare with are listed as follows:   	

	\begin{itemize} 
		\item \textit{ITTT}\cite{cai2014new}: An iterative thresholding method controlled by the distance between current and previous thresholds.  
		
		\item \textit{CAT}\cite{win2015contrast}: A modified Otsu's thresholding method \cite{otsu1979threshold} with the enhancement of contrast via resizing the histogram.   
	\end{itemize}

		\section{Results and Discussion}\label{discussion}
	
	\subsection{Comparison with results from different frameworks}
\begin{figure}[h!]%
	\centering
	\begin{subfigure}{0.48\columnwidth}
		\includegraphics[width=\columnwidth]{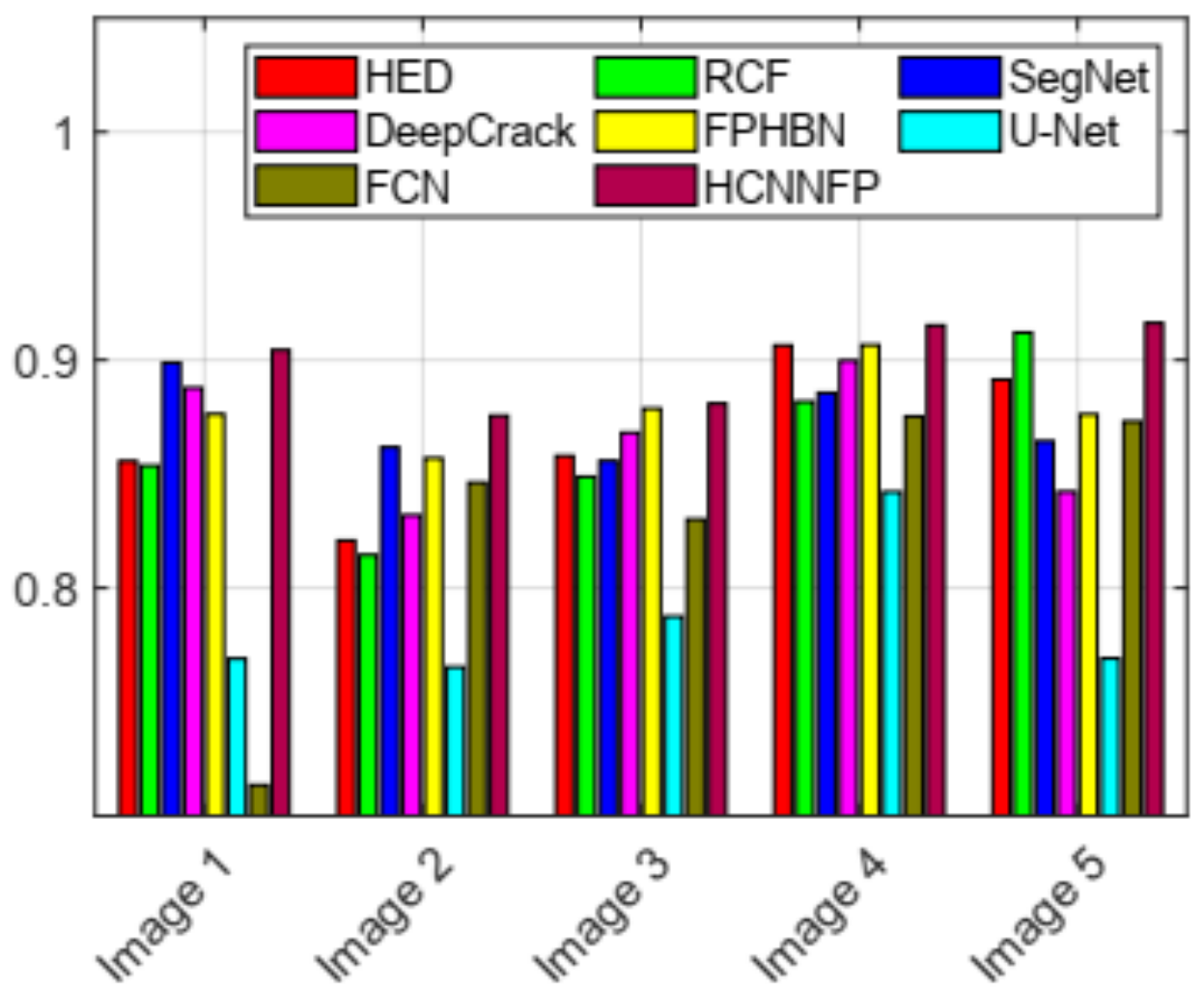}%
		\caption{}%
		\label{subfiga}%
	\end{subfigure}\hfill%
	\begin{subfigure}{0.48\columnwidth}
		\includegraphics[width=\columnwidth]{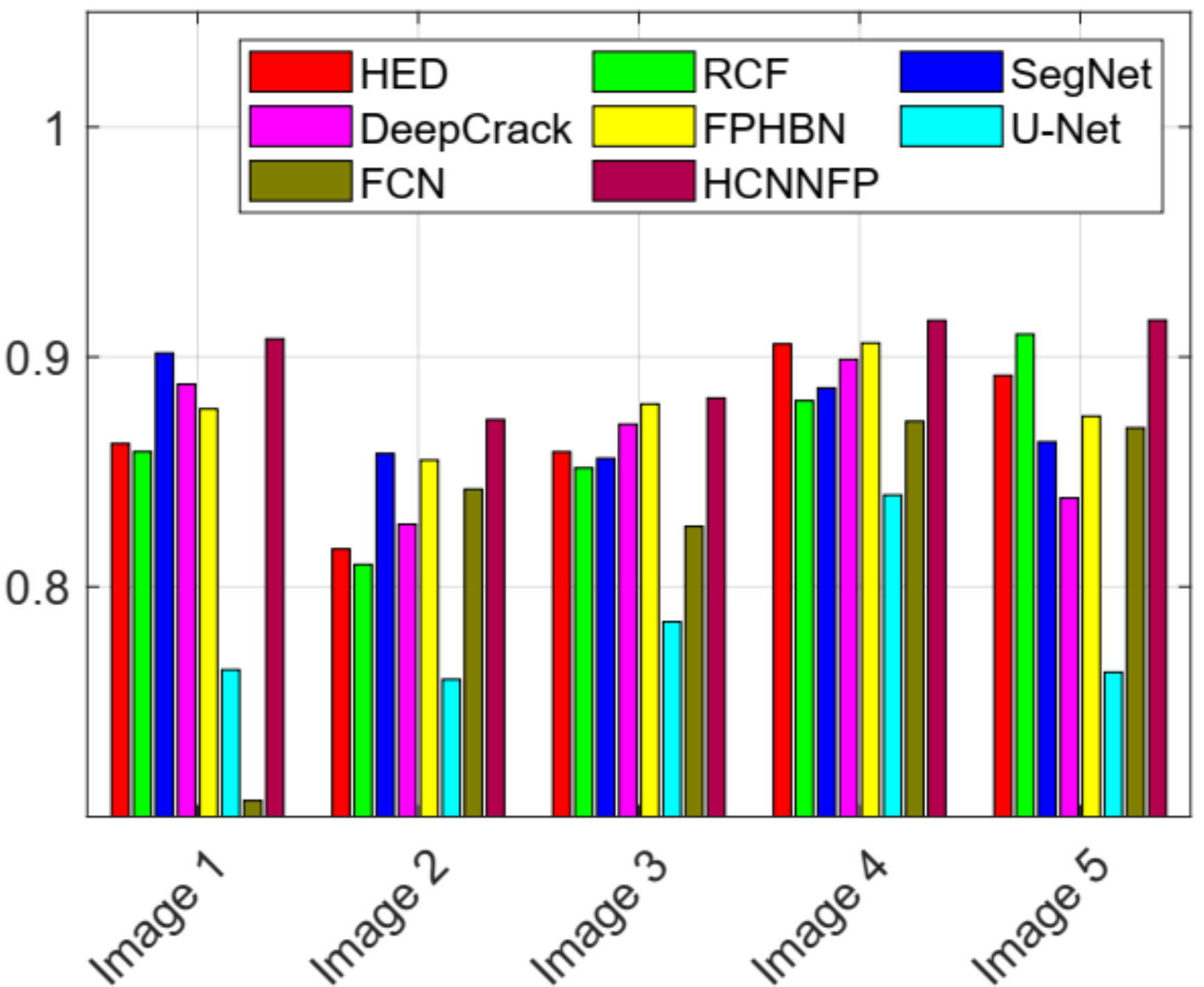}%
		\caption{}%
		\label{subfigc}%
	\end{subfigure}		\hfill  \\
	\begin{subfigure}{0.48\columnwidth}
		\includegraphics[width=\columnwidth]{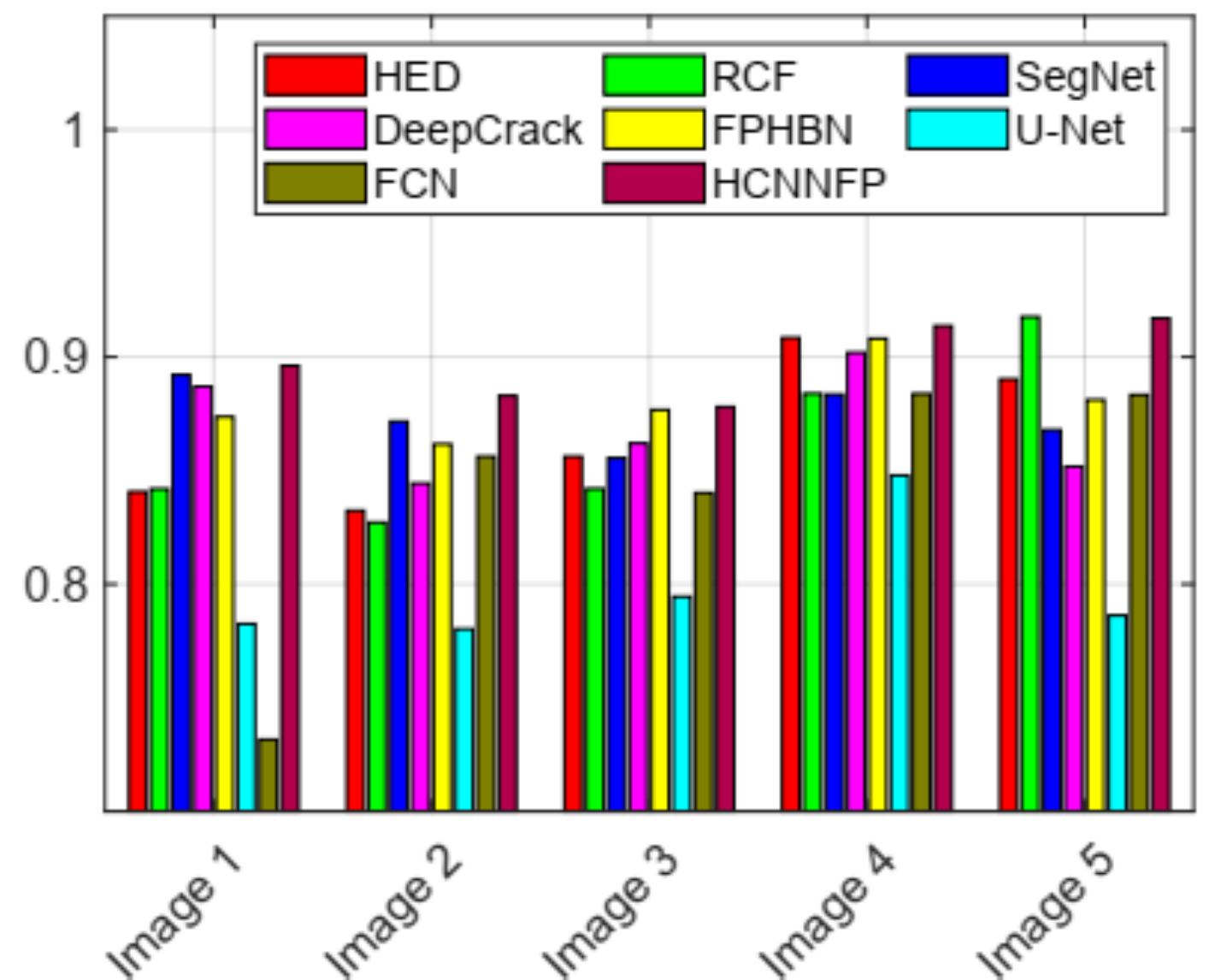}%
		\caption{}%
		\label{subfiga}%
	\end{subfigure}\hfill%
	\begin{subfigure}{0.50\columnwidth}
		\includegraphics[width=\columnwidth]{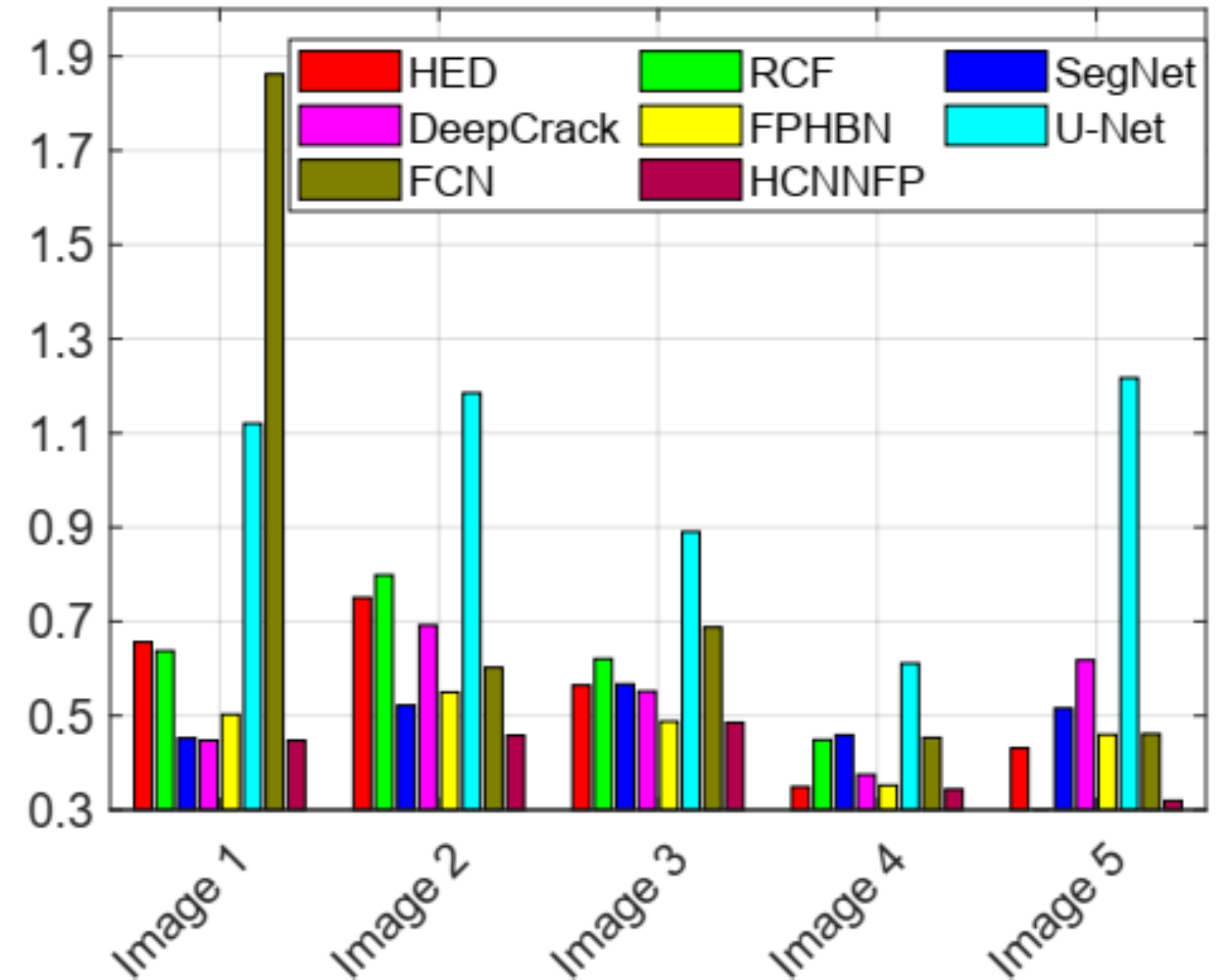}%
		\caption{}%
		\label{subfigc}%
	\end{subfigure}%
	\caption{Quantitative results of the first five samples on crack images: (a) $F_\beta|\beta^2=0.25$, (b) $F_\beta|\beta^2=0.3$, (c) $AF_\beta$, (d) $MAPE$.}\vspace{-0.15cm}
	\label{fig:DLC1}
\end{figure}

\begin{figure}[h!]%
	\centering
	\begin{subfigure}{0.48\columnwidth}
		\includegraphics[width=\columnwidth]{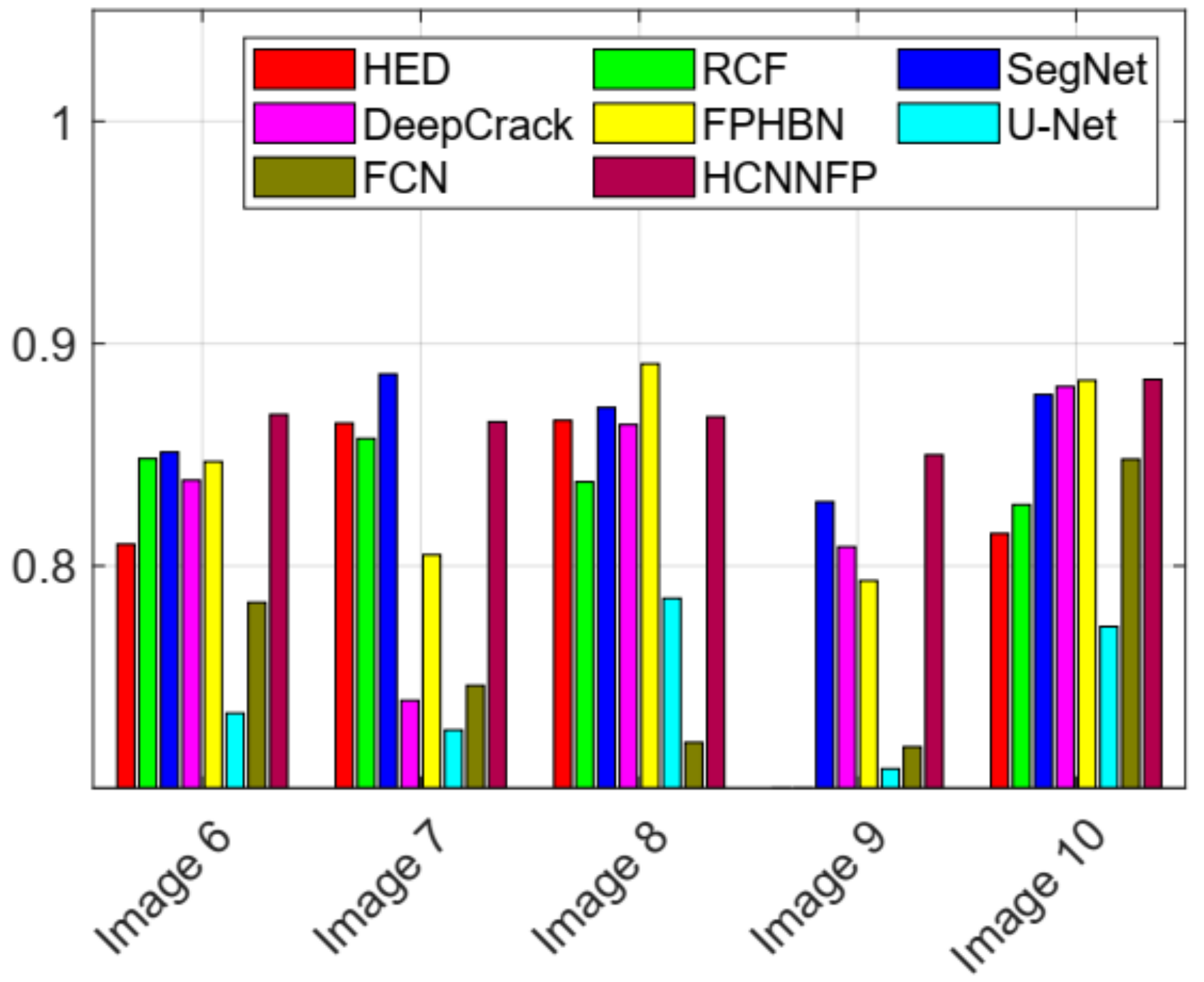}%
		\caption{}%
		\label{subfiga}%
	\end{subfigure}\hfill%
	\begin{subfigure}{0.48\columnwidth}
		\includegraphics[width=\columnwidth]{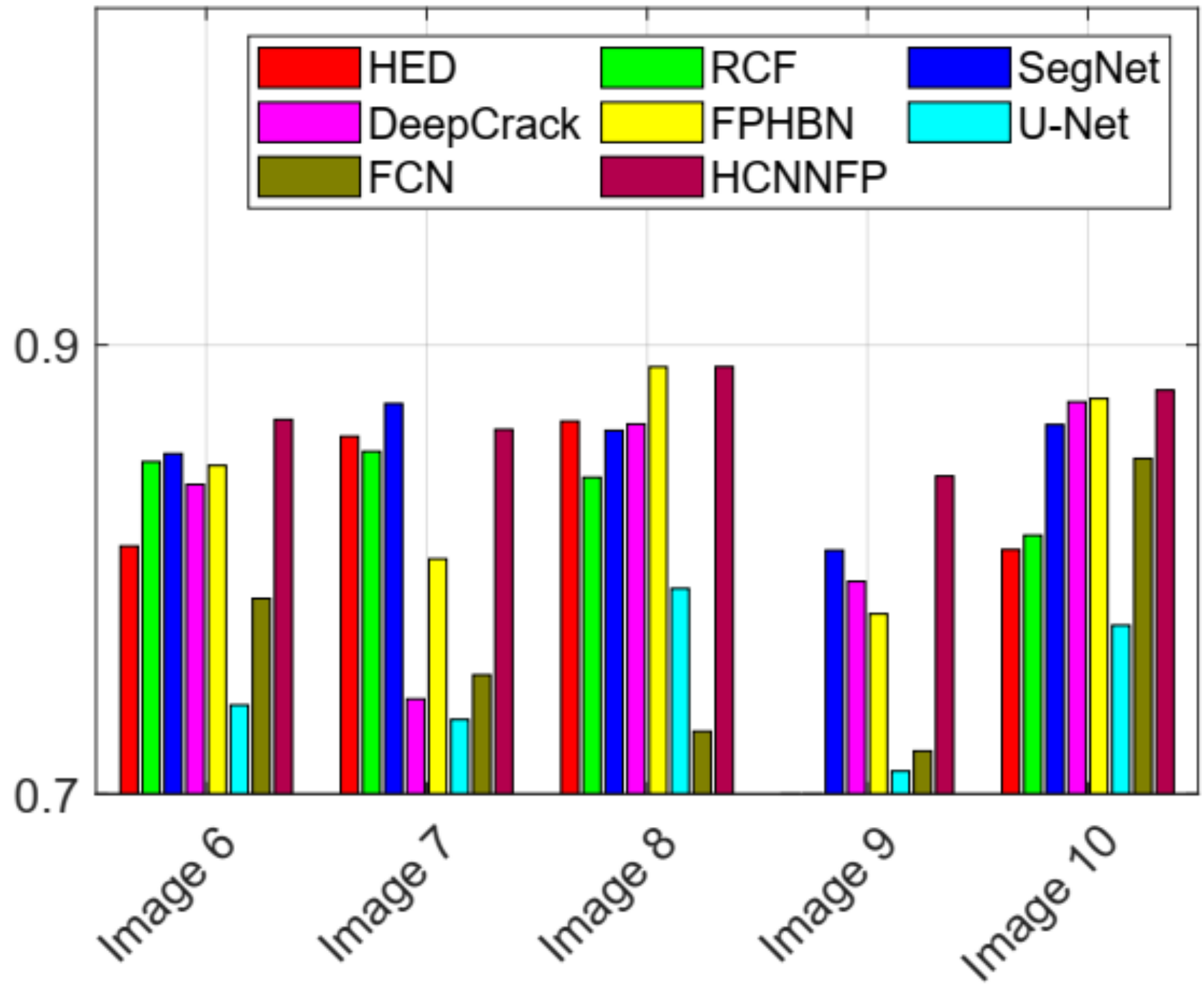}%
		\caption{}%
		\label{subfigc}%
	\end{subfigure}		\hfill  \\
	\begin{subfigure}{0.48\columnwidth}
		\includegraphics[width=\columnwidth]{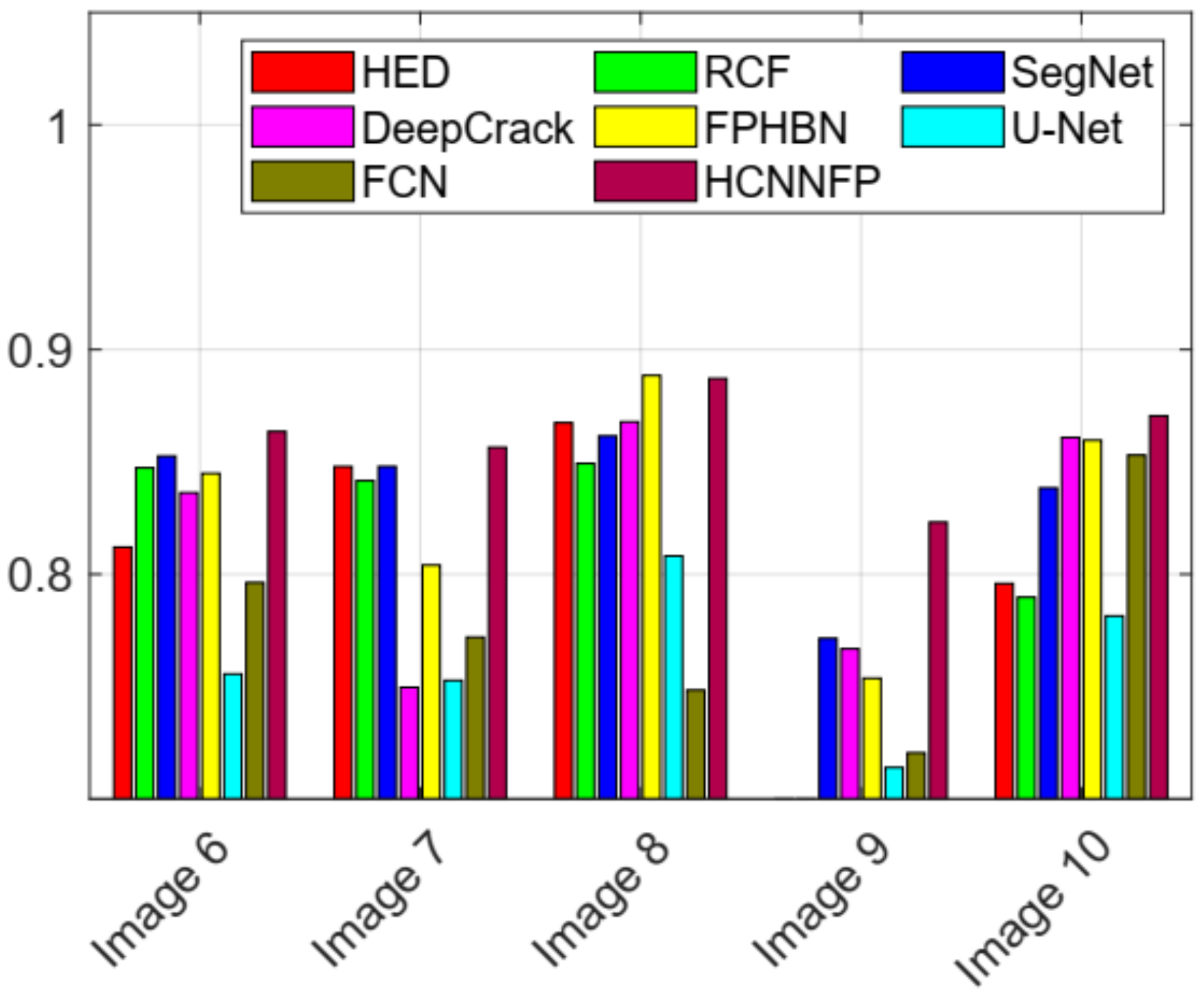}%
		\caption{}%
		\label{subfiga}%
	\end{subfigure}\hfill%
	\begin{subfigure}{0.50\columnwidth}
		\includegraphics[width=\columnwidth]{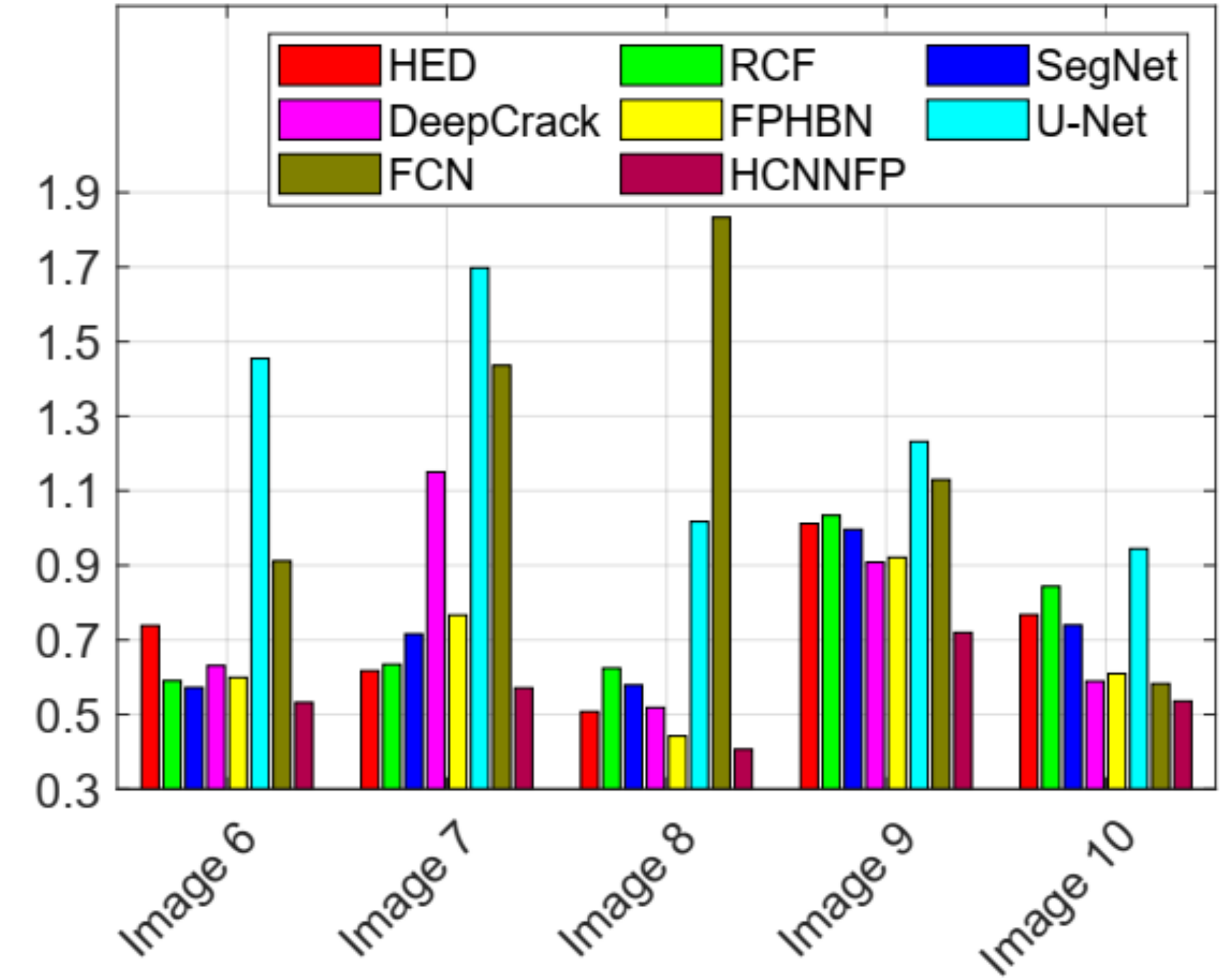}%
		\caption{}%
		\label{subfigc}%
	\end{subfigure}%
	\caption{Quantitative results of the last five samples on crack images: (a) $F_\beta|\beta^2=0.25$, (b) $F_\beta|\beta^2=0.3$, (c) $AF_\beta$, (d) $MAPE$.}\vspace{-0.15cm}
	\label{fig:DLC2}
\end{figure}

	The visual results of those DCNN frameworks for crack detection are depicted in Fig. \ref{fig:comparison1}, wherein autoencoder models such as SegNet, FPBHN, DeepCrack, and HCNNFP are capable of resisting the interference caused by texture, painted boundaries, and uneven lighting conditions. Notably, in addition effectively preserving completed contour of cracks better than other methods, including DeepCrack and FPHBN, the proposed HCNNFP is also able to remove those confusing patterns, as shown in the last five columns for DCD-7 dataset. The complexity of crack-like patterns actively contributes to a better F-measure with our method. Besides, although the results of DeepCrack and HCNNFP reach a similar level of sophistication, HCNNFP outputs a thinner outline of cracks. All these lead to less false-positive labeling around the contour and more accurate prediction.  
	This advantage can be confirmed quantitatively by the measures $AF_\beta$ and $MAPE$ as shown in the charts of Fig. \ref{fig:DLC1} and \ref{fig:DLC2}.
	
	For further comparison in the first experiment, the average measures for 6 DCNN approaches are listed in Table \ref{dlcomparison1} and Table \ref{dlcomparison2}. It can be seen that our HCNNFP obtains the highest $AF_\beta$ and the lowest $MAPE$ for CrackForest, SydCrack, DCD, and GAPs datasets, and performs as the second-best in Crack500. The processing time of all the compared models is demonstrated in Table \ref{dlcomparison3}. Since all the tested networks are fed with the same data for testing and data loader, the difference of processing time can be considered as a relative comparison of computational consumption. As shown in Table \ref{dlcomparison3},  among the top three approaches in terms of crack detection accuracy, namely HCNNFP, DeepCrack, and FPHBN, our proposed approach ranked second in computational efficiency. More importantly, its ability to process cropped images at approximately 60 frames per second has demonstrated the capability of our method in real-time application. It is noted that the augmentation method used for SYDcrack and CrackForest is by cropping rather than resizing as in Crack500. Since resizing can generally weaken the representation of features with fewer details, DCD and GAPs are used here in the original size from the source provided by \cite{liu2019deep} with a higher fidelity level.  
	
\begin{figure}[h!]
	\centering
	\includegraphics[width=1\columnwidth]{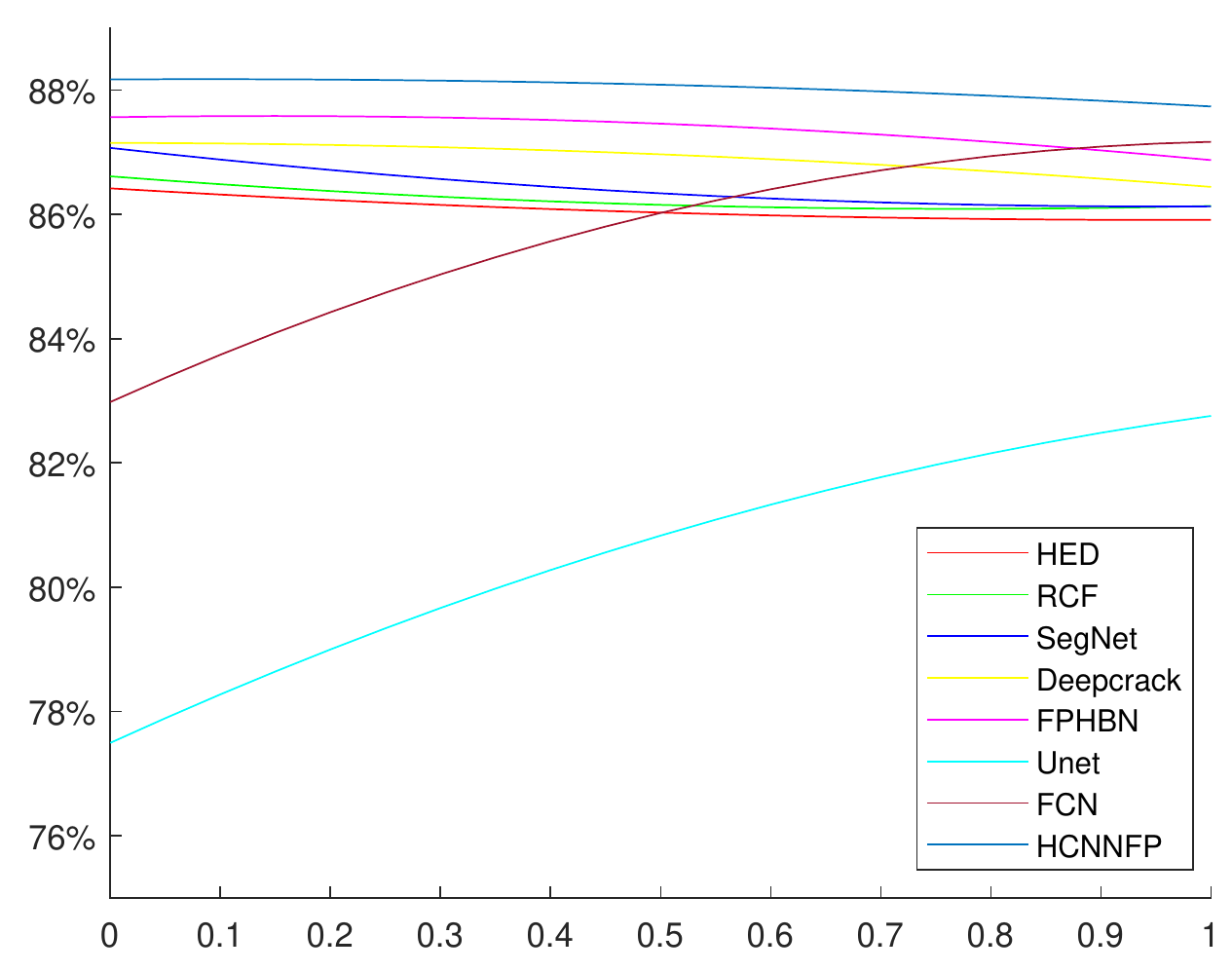}
	\caption{Distribution of $F_\beta$ with respect to $\beta^2$}
	\label{fig:betacurve1}
\end{figure}

	The feature preserving capability and high performance in crack detection as evaluated by those measures indicate the effectiveness of our approach overall. It is also worth noting that the top three models are all U-shape autoencoder while the fourth is also based on the first model architecture.  This indicates the advantage of the proposed feature preservation branch applied to existing hierarchical architectures for vision-based monitoring. Specifically, recent autoencoder models such as DeepCrack, FPHBN and our HCNNFP performs better than the prototype autoencoders like SegNet. The main difference between them is that those updated models has an independent branch to integrate the output from different scales into a unified scale after resizing and refilling. This branch can be the key to the improvement of accuracy. Notably, the robustness of our proposed method over the range of interest for value $\beta^2$. Indeed, the relationship of $F_\beta$ versus $\beta^2$ for the CrackForest dataset is shown in Fig. \ref{fig:betacurve1}, where it can be seen that the fitness $F_\beta$ of HCNNFP remains the highest for $\beta^2 \leq 1$ as compared to other existing deep learning techniques. In particular, the evaluation using the standard F-measure can also be seen in Fig. \ref{fig:betacurve1}, where, at the point $\beta=1$, the proposed HCNNFP gives the maximal value at around 0.88. This merit is also preserved if taking the arithmetic mean of $F_\beta$ for the five datasets.          
	
	\begin{table*}[tbh!]
	\centering
	\renewcommand\thetable{II}
	\begin{tabular}{p{1.1cm}|cccccccccc}
		\hline
		& \multicolumn{2}{c|}{Crack500}      & \multicolumn{2}{c|}{CrackForest}               & \multicolumn{2}{c|}{SYDCrack}     & \multicolumn{2}{c|}{DCD} & \multicolumn{2}{c}{GAPs}                                                       \\ \cline{2-11} 
		\multirow{-2}{*}{Methods} & 
		\multicolumn{1}{p{0.75cm}<{\centering}|}{$_{\beta^2=0.25}$}        & \multicolumn{1}{p{0.75cm}<{\centering}|}{$_{\beta^2=0.3}$}      &
		\multicolumn{1}{p{0.75cm}<{\centering}|}{$_{\beta^2=0.25}$}        & \multicolumn{1}{p{0.75cm}<{\centering}|}{$_{\beta^2=0.3}$}  	&
		\multicolumn{1}{p{0.75cm}<{\centering}|}{$_{\beta^2=0.25}$}        & \multicolumn{1}{p{0.75cm}<{\centering}|}{$_{\beta^2=0.3}$}  	& \multicolumn{1}{p{0.75cm}<{\centering}|}{$_{\beta^2=0.25}$}        & \multicolumn{1}{p{0.75cm}<{\centering}|}{$_{\beta^2=0.3}$}      & \multicolumn{1}{p{0.75cm}<{\centering}|}{$_{\beta^2=0.25}$}        & 
		$_{\beta^2=0.3}$                           \\ \hline
HED       & 0.7877      & 0.7868  & 0.8699     & 0.8697        & 0.8414 & 0.8417 &  0.8350&  0.8327& 0.6664 &	0.6626 \\
RCF       & 0.8034      & 0.8024  & 0.8612     & 0.8602        & 0.8462 & 0.8473 & 0.8542 &  0.8509& 0.6568 &  0.6515 \\
SegNet    & 0.8024      & 0.8021  & 0.8654     & 0.8642        & 0.8507 & 0.8509 & 0.8588 &  0.8570 &0.7536 &	0.7530\\
DeepCrack & 0.8068      & 0.8083  & 0.8765     & 0.8765       & 0.8514 & 0.8514 &     0.8667& 0.8656 &0.7642 &	0.7657\\
FPHBN     & \textbf{0.8211}       & \textbf{0.8207} & 0.8773     & 0.8771        & 0.8510 & 0.8517  &     0.8671& 0.8653& 0.7890	&0.7816      \\
U-Net      & 0.7510      & 0.7542  & 0.7839     & 0.7853        & 0.7730 & 0.7778 &    0.7849 & 0.7897& 0.6770 & 0.6831 \\
FCN       & 0.8030      & 0.8042  & 0.8417     & 0.8452         & 0.8243 & 0.8274 & 		0.8005 &0.8049 & 0.7562	& 0.7619  \\
HCNNFP     & 0.8179      & 0.8181  & \textbf{0.8805}     & \textbf{0.8797}        & \textbf{0.8552} & \textbf{0.8551} & 	\textbf{0.8700} &\textbf{0.8688}	&\textbf{0.7949}&	\textbf{0.7933}       \\ \hline
	\end{tabular}
	\caption{Comparison of F-measure $F_\beta$ among eight DCNN approaches on five datasets.}
	\label{dlcomparison1}
\end{table*}

	\begin{table*}[tbh!]
		\centering
			\renewcommand\thetable{III}
		\begin{tabular}{p{1.1cm}|cccccccccp{0.85cm}<{\centering}}
			\hline
			& \multicolumn{2}{c|}{Crack500}      & \multicolumn{2}{c|}{CrackForest}                              & \multicolumn{2}{c|}{SYDCrack}  & \multicolumn{2}{c|}{DCD} & \multicolumn{2}{c}{GAPs}                                                                   \\ \cline{2-11} 
			\multirow{-2}{*}{Methods} & \multicolumn{1}{p{0.75cm}<{\centering}|}{$AF_\beta$}        & \multicolumn{1}{p{0.75cm}<{\centering}|}{$MAPE$}      & \multicolumn{1}{p{0.75cm}<{\centering}|}{$AF_\beta$}        & \multicolumn{1}{p{0.75cm}<{\centering}|}{$MAPE$}      & 
			\multicolumn{1}{p{0.75cm}<{\centering}|}{$AF_\beta$}        & \multicolumn{1}{p{0.75cm}<{\centering}|}{$MAPE$}      & \multicolumn{1}{p{0.75cm}<{\centering}|}{$AF_\beta$}        & \multicolumn{1}{p{0.75cm}<{\centering}|}{$MAPE$}      & 
			\multicolumn{1}{p{0.75cm}<{\centering}|}{$AF_\beta$}        & $MAPE$                           \\ \hline
HED       & 0.7849      & 0.9038 & 0.8692     & 0.5095       & 0.8432     & 0.7358 &  0.8277 & 0.9856& 0.6542  &1.2491\\
RCF       & 0.8005      & 0.8325 & 0.8579     & 0.5498       & 0.8508     & 0.7360 &0.8434 &0.9558 &0.6511 & 1.1798\\
SegNet    & 0.8016      & 0.9619 & 0.8616     & 0.5430       & 0.8523     & 0.7156 &0.8505 &0.8602 &0.7322 & 1.2090\\
DeepCrack & 0.8122      & 0.8946 & 0.8767     & 0.4870       & 0.8522     & 0.6994  & 0.8632 & 0.7649 &0.7538 &	0.9617     \\
FPHBN     & \textbf{0.8202}      & \textbf{0.7782}& 0.8768     & 0.4843         & 0.8542     & 0.6863 &0.8615 & 0.7604 & 0.7662	&0.8723   \\
U-Net      & 0.7629      & 1.3783 & 0.7887     & 0.8330      & 0.7907     & 1.2924   &0.8062 & 1.6089  &0.6927 & 1.8614 \\
FCN       & 0.8076      & 0.9251 & 0.8542     & 0.6225      & 0.8358     & 0.8500  & 0.8170 & 1.4950 & 0.7685 & 1.1841  \\
HCNNFP     & 0.8188      & 0.8081 & \textbf{0.8780}     &\textbf{0.4807}         & \textbf{0.8558} & \textbf{0.6725}&\textbf{0.8662} & \textbf{0.7520} & \textbf{0.7807}  & \textbf{0.8503}         \\ \hline
		\end{tabular}
		\caption{Comparison of average measures among eight DCNN approaches on five datasets.}
		\label{dlcomparison2}
	\end{table*}

	\begin{table*}[tbh!]
	\renewcommand{\arraystretch}{1.3}
	\footnotesize\addtolength{\tabcolsep}{-5pt}
	\begin{center}
		\begin{tabular}{cllll}	
			Crack500-11  &	
			\begin{subfigure}{0.12\textwidth}\centering\includegraphics[width=\linewidth]{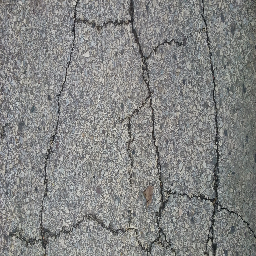}\label{fig:taba}\end{subfigure}
			\begin{subfigure}{0.12\textwidth}\centering\includegraphics[width=\linewidth]{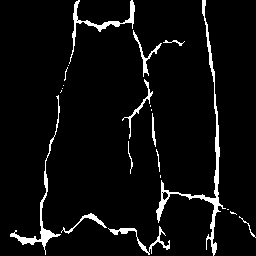}\label{fig:taba}\end{subfigure}
			\begin{subfigure}{0.12\textwidth}\centering\includegraphics[width=\linewidth]{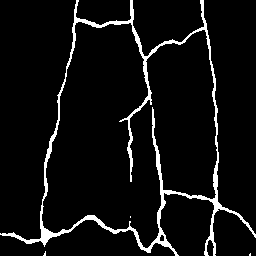}\label{fig:tabc}\end{subfigure}
			\begin{subfigure}{0.12\textwidth}\centering\includegraphics[width=\linewidth]{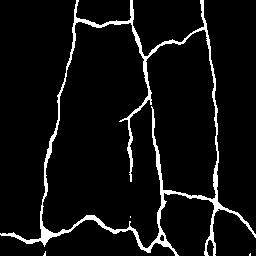}\label{fig:tabc}\end{subfigure}
			\begin{subfigure}{0.12\textwidth}\centering\includegraphics[width=\linewidth]{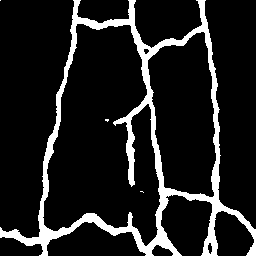}\label{fig:tabc}\end{subfigure}					
			\begin{subfigure}{0.12\textwidth}\centering\includegraphics[width=\linewidth]{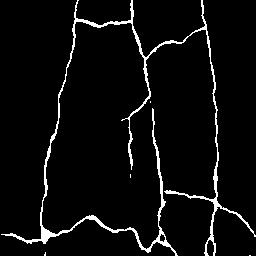}\label{fig:tabc}\end{subfigure}\vspace{3px}\\[3mm]	
			\small{CrackForest-12}  &	
			\begin{subfigure}{0.12\textwidth}\centering\includegraphics[width=\linewidth]{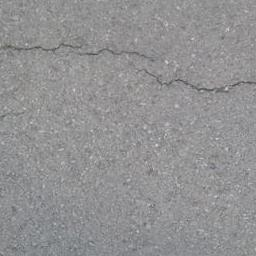}\label{fig:taba}\end{subfigure}
			\begin{subfigure}{0.12\textwidth}\centering\includegraphics[width=\linewidth]{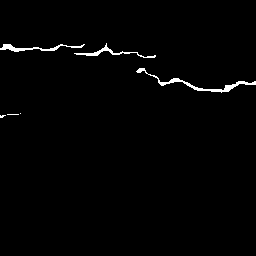}\label{fig:taba}\end{subfigure}
			\begin{subfigure}{0.12\textwidth}\centering\includegraphics[width=\linewidth]{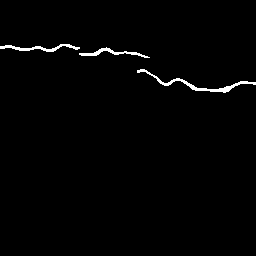}\label{fig:tabc}\end{subfigure}
			\begin{subfigure}{0.12\textwidth}\centering\includegraphics[width=\linewidth]{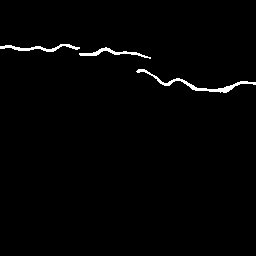}\label{fig:tabc}\end{subfigure}
			\begin{subfigure}{0.12\textwidth}\centering\includegraphics[width=\linewidth]{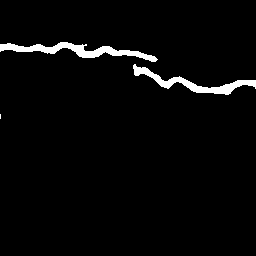}\label{fig:tabc}\end{subfigure}				
			\begin{subfigure}{0.12\textwidth}\centering\includegraphics[width=\linewidth]{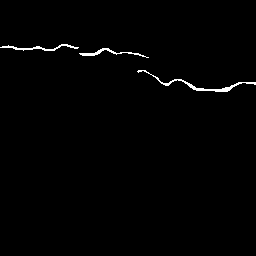}\label{fig:tabc}\end{subfigure}\vspace{3px}\\[3mm]
			SYDCrack-13  &		
			\begin{subfigure}{0.12\textwidth}\centering\includegraphics[width=\linewidth]{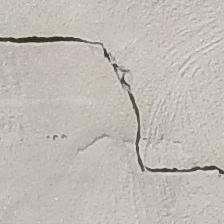}\label{fig:taba}\end{subfigure}
			\begin{subfigure}{0.12\textwidth}\centering\includegraphics[width=\linewidth]{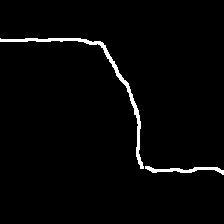}\label{fig:taba}\end{subfigure}
			\begin{subfigure}{0.12\textwidth}\centering\includegraphics[width=\linewidth]{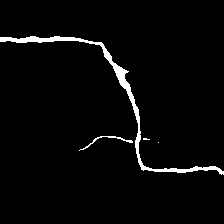}\label{fig:tabc}\end{subfigure}
			\begin{subfigure}{0.12\textwidth}\centering\includegraphics[width=\linewidth]{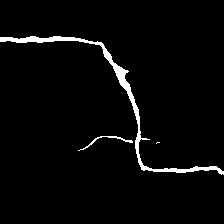}\label{fig:tabc}\end{subfigure}
			\begin{subfigure}{0.12\textwidth}\centering\includegraphics[width=\linewidth]{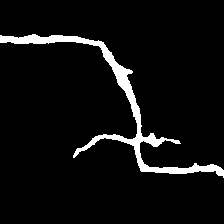}\label{fig:tabc}\end{subfigure}				
			\begin{subfigure}{0.12\textwidth}\centering\includegraphics[width=\linewidth]{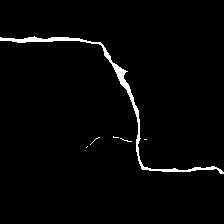}\label{fig:tabc}\end{subfigure}\vspace{3px}\\
			&\hspace{0.032\textwidth}Original\hspace{0.088\textwidth}GT\hspace{0.084\textwidth}Fixed 0.5\hspace{0.078\textwidth}ITTT\hspace{0.088\textwidth}CAT\hspace{0.088\textwidth}CBAT\hspace{0.032\textwidth}
			
		\end{tabular}
		\captionof{figure}{Binarization results of the probability map.}\vspace{-0.35cm}
		\label{fig:comparison2}
	\end{center}
\end{table*}

	\begin{table*}[tbh!]
	\centering
	\renewcommand\thetable{IV}
	\begin{tabular}{p{2.6cm}|cccccccc}
		\hline

		Methods     &      	HED      & 	RCF   & 	SegNet  & 	DeepCrack  & FPHBN & U-Net & FCN & HCNNFP  \\ \hline
		Processing time(ms) 	      & 7.44 & 9.47  & 14.49 & 15.24  & 18.15       & 15.12 & 12.71   & 16.01     \\ \hline
	\end{tabular}
	\caption{Comparison of the processing time among eight DCNN approaches.}
	\label{dlcomparison3}
\end{table*}

	\subsection{Comparison in post-processing}
\begin{figure}[tbh!]%
		\centering
		\begin{subfigure}{0.48\columnwidth}
			\includegraphics[width=\columnwidth]{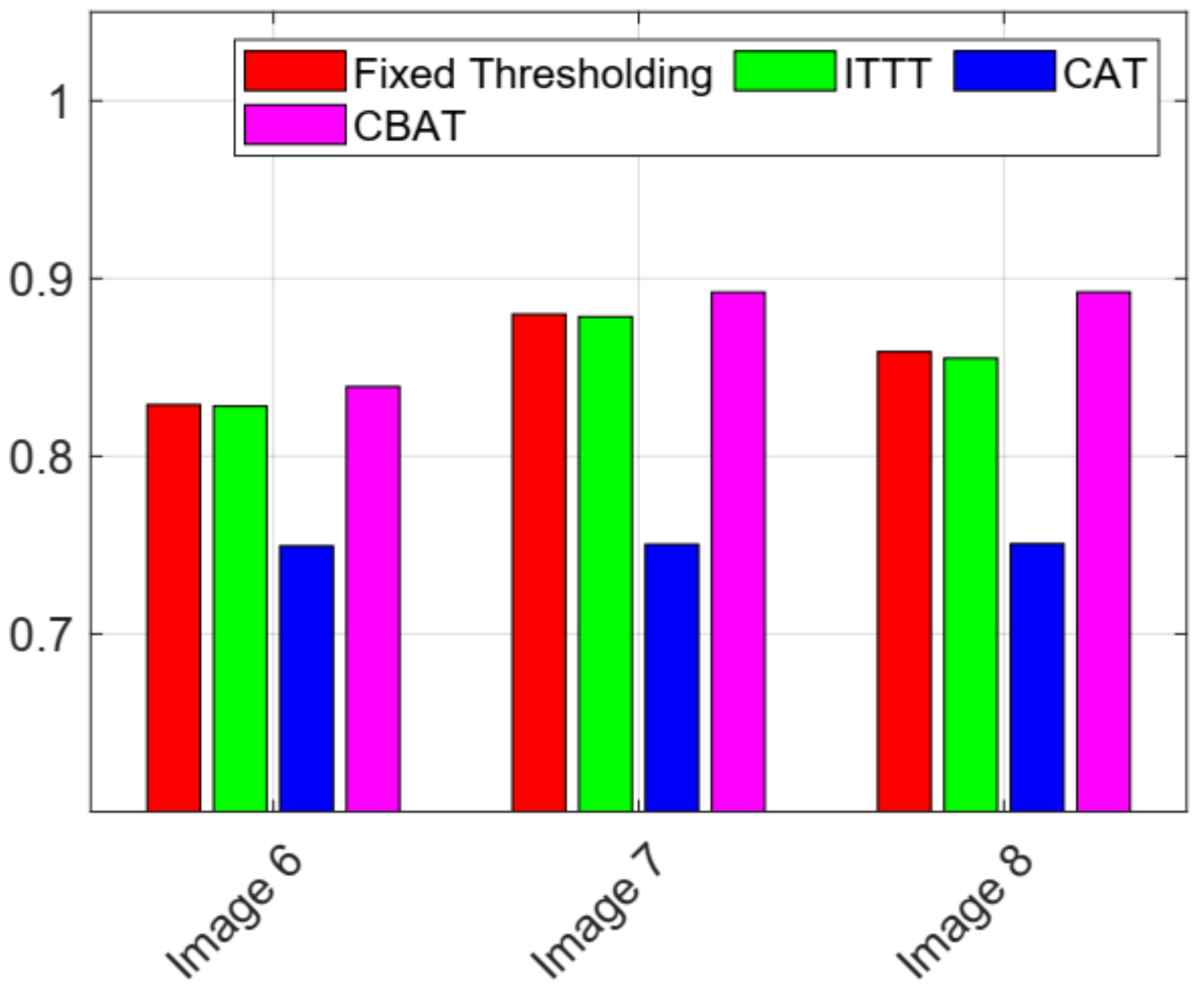}%
			\caption{}%
			\label{subfiga}%
		\end{subfigure}\hfill%
		\begin{subfigure}{0.48\columnwidth}
			\includegraphics[width=\columnwidth]{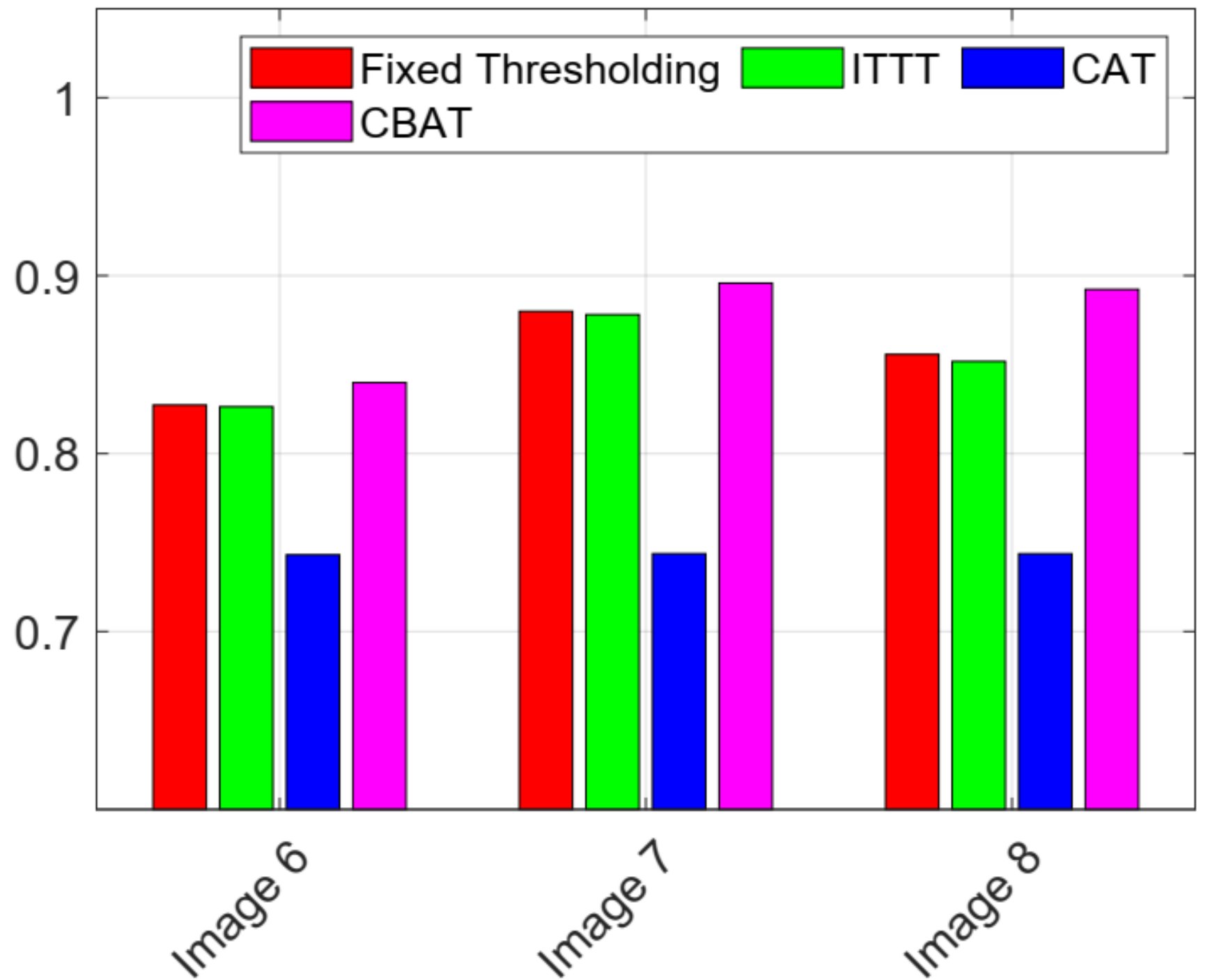}%
			\caption{}%
			\label{subfigc}%
		\end{subfigure}		\hfill  \\	
				\begin{subfigure}{0.48\columnwidth}
					\includegraphics[width=\columnwidth]{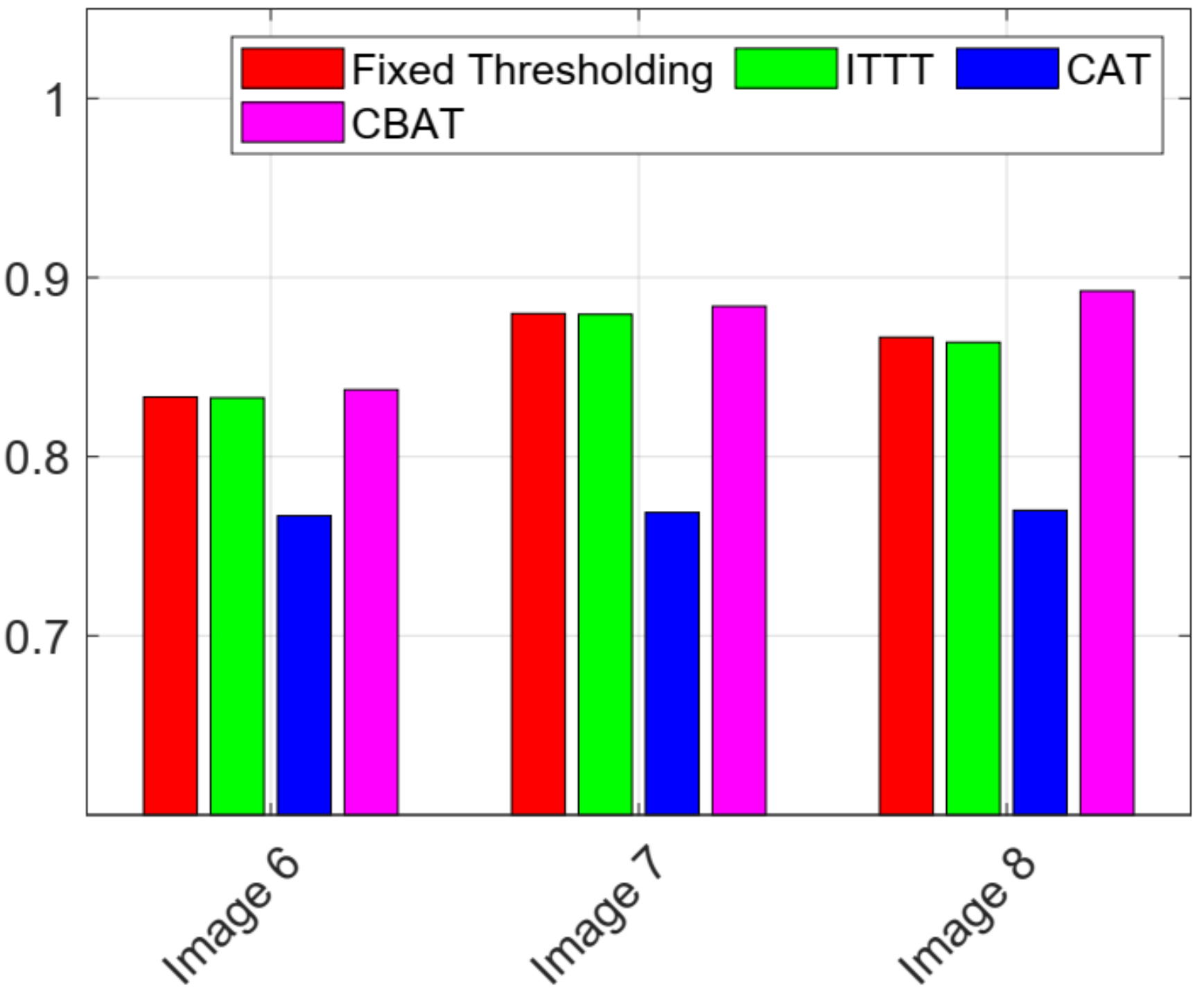}%
					\caption{}%
					\label{subfiga}%
				\end{subfigure}\hfill%
				\begin{subfigure}{0.48\columnwidth}
					\includegraphics[width=\columnwidth]{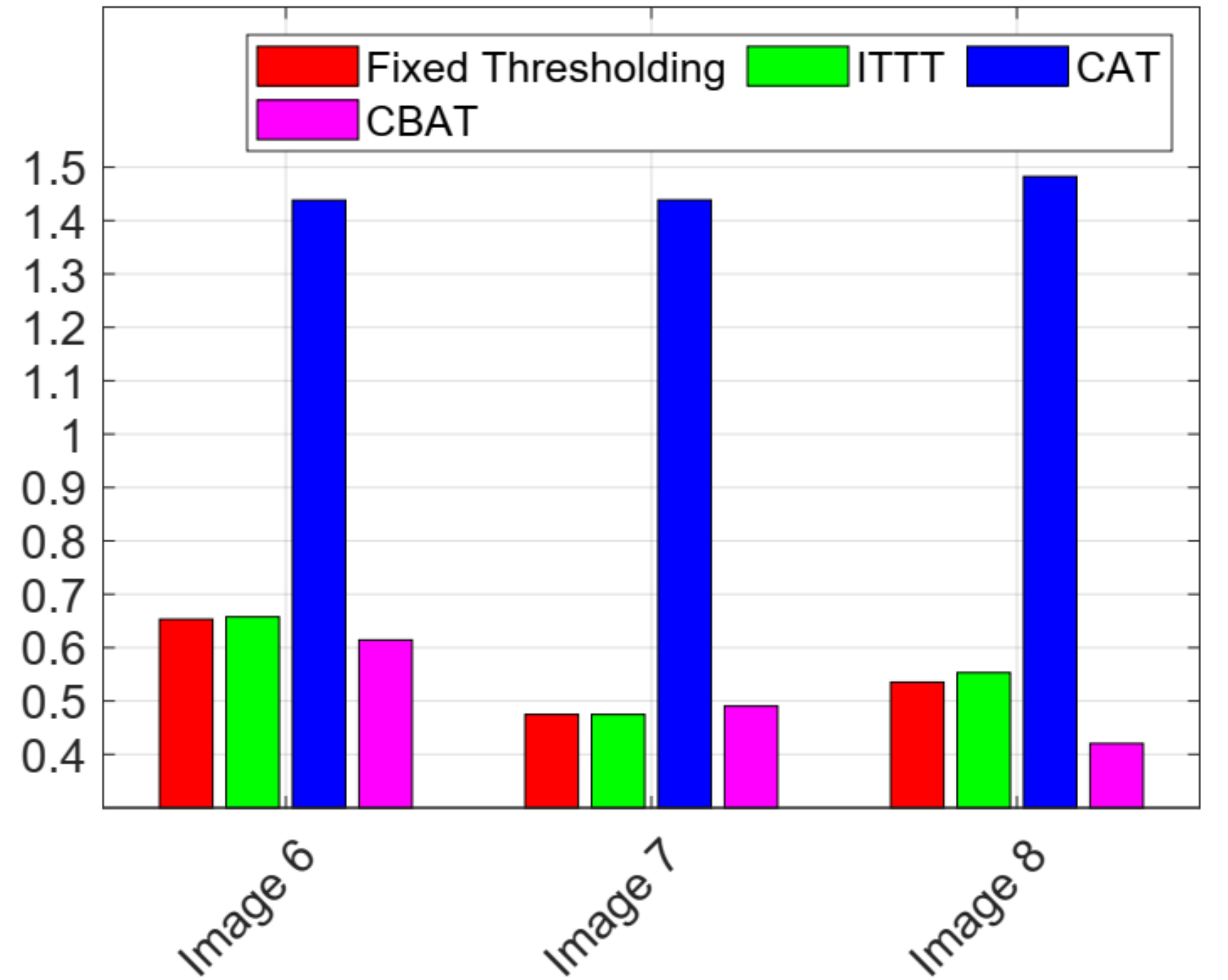}%
					\caption{}%
					\label{subfigc}%
		\end{subfigure}
	\caption{Quantitative results of fixed thresholding and CBAT on crack images: (a) $F_\beta|\beta^2=0.25$, (b) $F_\beta|\beta^2=0.3$, (c) $AF_\beta$, (d) $MAPE$.}\vspace{-0.15cm}		
		\label{fig:POST}
	\end{figure}

	The binarization results are shown in Fig. \ref{fig:comparison2}. It can be seen that all the approaches can provide a high level of fitness to the ground truth. However, among them, CBAT presents a prediction map with the fewest crack labels. Moreover, as shown in the second row, although both thresholding methods are misled by the trace of insignificant dents, our CBAT can reduce the error by removing some false-positive pixels, and thus enhancing the precision rate. This improvement is explained by the high credibility of CBAT in favor of crack features, and by its low credibility in shadow dents due to fewer rounds in the repetition of similar patterns in the training set. As such patterns are excluded by using CBAT, the prediction is closer to the ground truth. This has resulted in the highest values of $AF_\beta$ and the lowest $MAE$ as shown in the chart of Fig. \ref{fig:POST}. The quantitative results of the four binarization approaches are presented in Table \ref{post1} and Table \ref{post2} for the three data sets CrackForest, SYDCrack and Crack500. Among the tests on three datasets, almost all metrics are better after applying CBAT to the post-processing step. 
	
	\begin{table}[tbh!]
	\centering
	\renewcommand\thetable{V}
	\begin{tabular}{p{1.1cm}|cccccp{0.85cm}<{\centering}}
		\hline
		& \multicolumn{2}{c|}{Crack500}      & \multicolumn{2}{c|}{CrackForest}                              & \multicolumn{2}{c}{SYDCrack}                                                           \\ \cline{2-7} 
\multirow{-2}{*}{Thresholds} & \multicolumn{1}{p{0.75cm}<{\centering}|}{$_{\beta=0.25}$}        & \multicolumn{1}{p{0.75cm}<{\centering}|}{$_{\beta=0.3}$}      & \multicolumn{1}{p{0.75cm}<{\centering}|}{$_{\beta=0.25}$}        & \multicolumn{1}{p{0.75cm}<{\centering}|}{$_{\beta=0.3}$}      & \multicolumn{1}{p{0.75cm}<{\centering}|}{$_{\beta=0.25}$}        & $_{\beta=0.3}$                      \\ \hline
		Fixed 0.5     & 0.8179      & 0.8181     & 0.8805     & 0.8797    & 0.8552     & 0.8551     \\
		ITTT          & 0.8167      & 0.8171     & 0.8789     & 0.8785    & 0.8541     & 0.8543     \\
		CAT           & 0.7456      & 0.7506     & 0.7777     & 0.7837    & 0.8007     & 0.8054     \\
		CBAT           & \textbf{0.8279}      & \textbf{0.8258}     &\textbf{ 0.8865}     & \textbf{0.8858}    & \textbf{0.8643}     &\textbf{ 0.8619 }	\\ \hline
	\end{tabular}
	\caption{Comparison of F-measure $F_\beta$ among binarization approaches: thresholding with fixed $T$=0.5, ITTT, CAT and CBAT.}
	\label{post1}
\end{table}	
	
	\begin{table}[tbh!]
		\centering
       \renewcommand\thetable{VI}
		\begin{tabular}{p{1.1cm}|cccccc}
			\hline
			\multirow{2}{*}{Thresholds} & \multicolumn{2}{c|}{Crack500} & \multicolumn{2}{c|}{CrackForest} & \multicolumn{2}{c}{SYDCrack}                      \\ \cline{2-7} 
			& \multicolumn{1}{p{0.75cm}<{\centering}|}{$AF_\beta$} & \multicolumn{1}{p{0.75cm}<{\centering}|}{$MAPE$} & \multicolumn{1}{p{0.75cm}<{\centering}|}{$AF_\beta$} & \multicolumn{1}{p{0.75cm}<{\centering}|}{$MAPE$} & \multicolumn{1}{p{0.75cm}<{\centering}|}{$AF_\beta$}  & $MAPE$ \\ \hline
			Fixed 0.5 & 0.8188      & 0.8081     & 0.8780     & 0.4807    & 0.8558     & 0.6725         \\
			ITTT      & 0.8183      & 0.8188     & 0.8777     & 0.4855    & 0.8554     & 0.6790         \\
			CAT       & 0.7642      & 1.6535     & 0.8001     & 1.1365    & 0.8181     & 1.1509         \\
			CBAT       & \textbf{0.8211}      & \textbf{0.7431}     & \textbf{0.8836}     & \textbf{0.4757}    & \textbf{0.8569}     & \textbf{0.6366}         \\
			\hline
		\end{tabular}
		\caption{Comparison of average measures among binarization approaches: thresholding with fixed $T$=0.5, ITTT, CAT and CBAT}
		\label{post2}
	\end{table}

	Notably, for an ablation analysis, in addition to the comparison of DeepCrack and HCNNFP for the cases without and with our feature preserving branch as shown in Fig. \ref{fig:comparison1} in the case of even binarization, the effect of autotuned thresholding is also presented in this comparison with our proposed CBAT. Indeed, as indicated in Table \ref{dlcomparison2}, the $AF_\beta$ are improved on all the datasets, especially on GAPs, with an increase of 2.69\%, while $MAPE$ drops by 11.14\%. With the proposed feature preserving branch, more false-negative samples are rectified due to its robust mechanism in dealing with the nonlinearity. Also, a similar improvement can be seen in the comparison between our autotuned thresholding and raw binarization as quantified in Table \ref{post2}. Those results verify the effectiveness and robustness of the proposed HCNNFP with feature preserving and autotuned thresholding.

	\subsection{Discussion}
	Experimental results have demonstrated performance enhancements from the proposed hierarchical convolutional neural network with feature preserving for crack detection towards intelligent monitoring of transportation infrastructure. In the post-processing stage, the proposed intercontrast iterative thresholding also significantly contributes to improving binarization results for accurate feature extraction. Experimental results in crack detection on different datasets have shown the influences of redundant nonlinearity on the level of detail abstraction and the need for high credibility and scalability for reliable assessment of the surface defects and its attributes. These issues can be effectively dealt with by using the proposed feature preserving branch and intercontrast iterative thresholding algorithm. Moreover, errors in vision-based defect detection are often not fully reflected by the currently used evaluation with $F_\beta$. As indicated in Fig. \ref{fig:comparison1}, the shift in evaluation results is rather small where some parts of features may be missing. Here, a more comprehensive metric like the average F-measure $AF_\beta$ offers a complementary criterion to consider also the effect of mislabeling due to unmatched labels. Future work will look at the incorporation of more information in post-processing with geometric filters to accommodate different shapes when classifying the probability with high credibility.

\section{Conclusion}\label{conclusion}
This paper has presented a hybrid framework for detection of surface cracks in roads, tunnels or bridges. The proposed hierarchical convolutional neural network is equipped with a feature preserving branch to deal with the trade-off between nonlinearity and information loss. Moreover, the credibility of the features at the network output is further improved with a new intercontrast iterative algorithm based on Otsu thresholding to increase the detection accuracy. From the raw prediction, our enhanced hierarchical neural network can alleviate deviations caused by nonlinearity accumulated along with the network depth such that the upper-layer features become more linear and worth more weighting in labeling. At the post-processing stage, the contrast-based iterative thresholding can automatically search for a suitable boundary value in the probability map for accurate binarization, subject to a robust $AF_\beta$ over a range of weighting between prediction and recall. As a result, the developed framework can successfully detect surface cracks of five different datasets for a road, a pavement, and a bridge subject to various texture levels. Extensive comparisons with the existing state-of-the-art deep learning convolutional neural networks for crack detection has demonstrated the merits of the proposed approach.

\bibliographystyle{IEEEtran}

\bibliography{mybibfile_ACCESS1}
\nocite{pan2020new}

%

\begin{IEEEbiography}[{\includegraphics[width=1in,height=1.25in,clip]{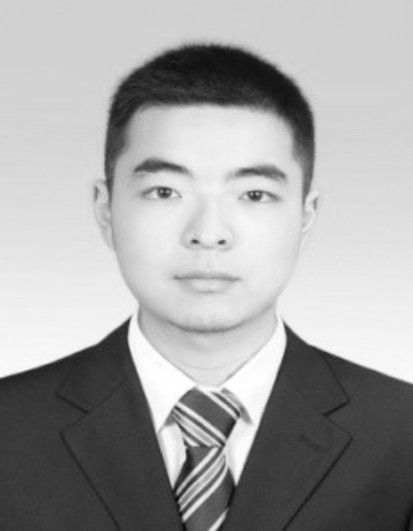}}]{Qiuchen Zhu} received the M.Eng. degree from Huazhong University of Science and Technology, Wuhan, China, in 2017.
	
	He is currently pursuing the Ph.D. degree with the School of Electrical and Data Engineering, University of Technology Sydney, Australia. His research interests include machine vision, image processing, probabilistic representation and uncertainty of deep learning. 
	
\end{IEEEbiography}

\begin{IEEEbiography}[{\includegraphics[width=1in,height=1.25in,clip]{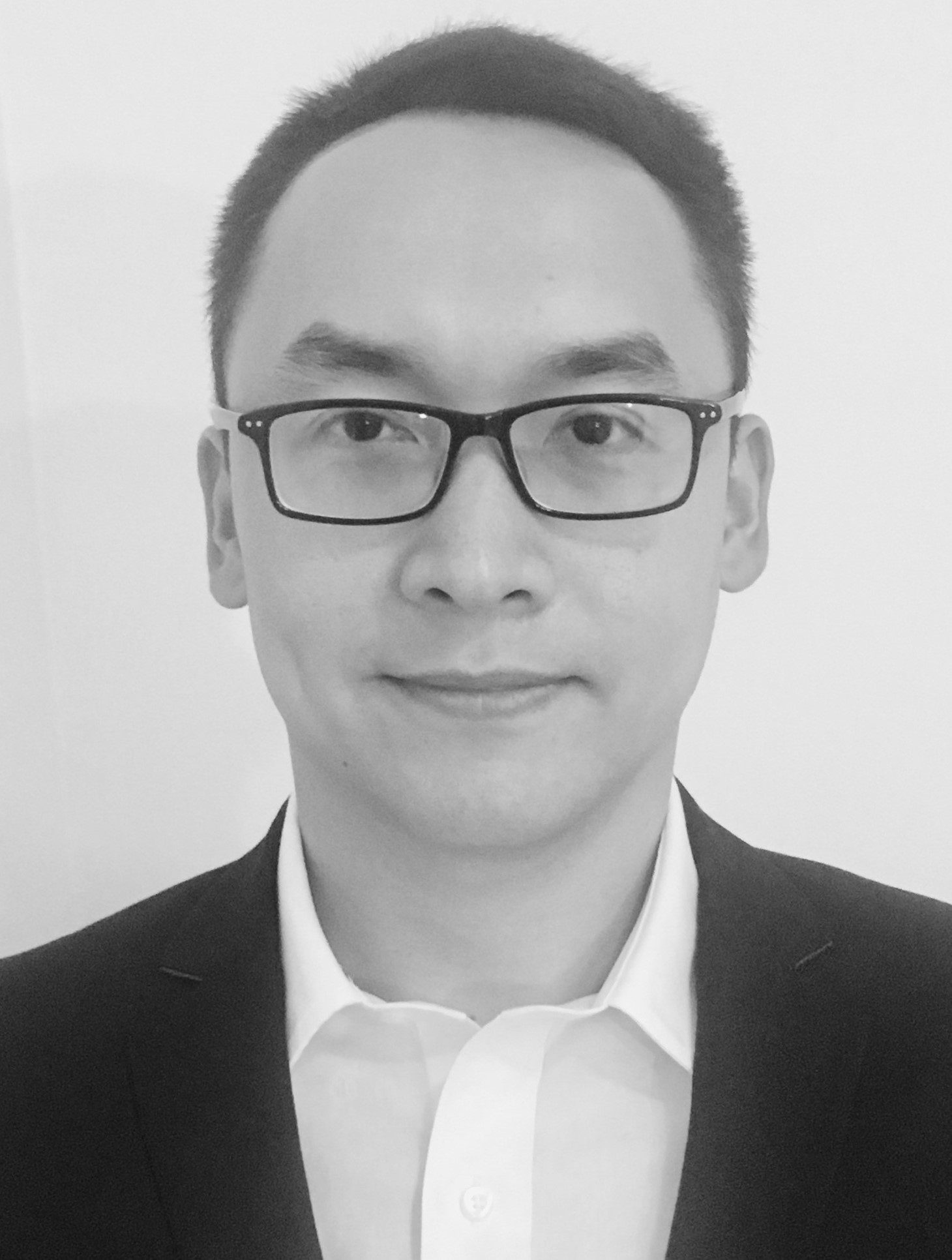}}]{Tran Hiep Dinh} received his M.Sc. degree in mechatronics from the Leibniz University Hanover, Germany, and PhD degree in engineering from the University of Technology Sydney, Australia, in 2010 and 2020, respectively.
	
	He is currently with the Faculty of Engineering Mechanics and Automation, VNU University of Engineering and Technology. His research interests include image processing, robotics, and machine learning.
	
\end{IEEEbiography}

\begin{IEEEbiography}[{\includegraphics[width=1in,height=1.25in,clip,keepaspectratio]{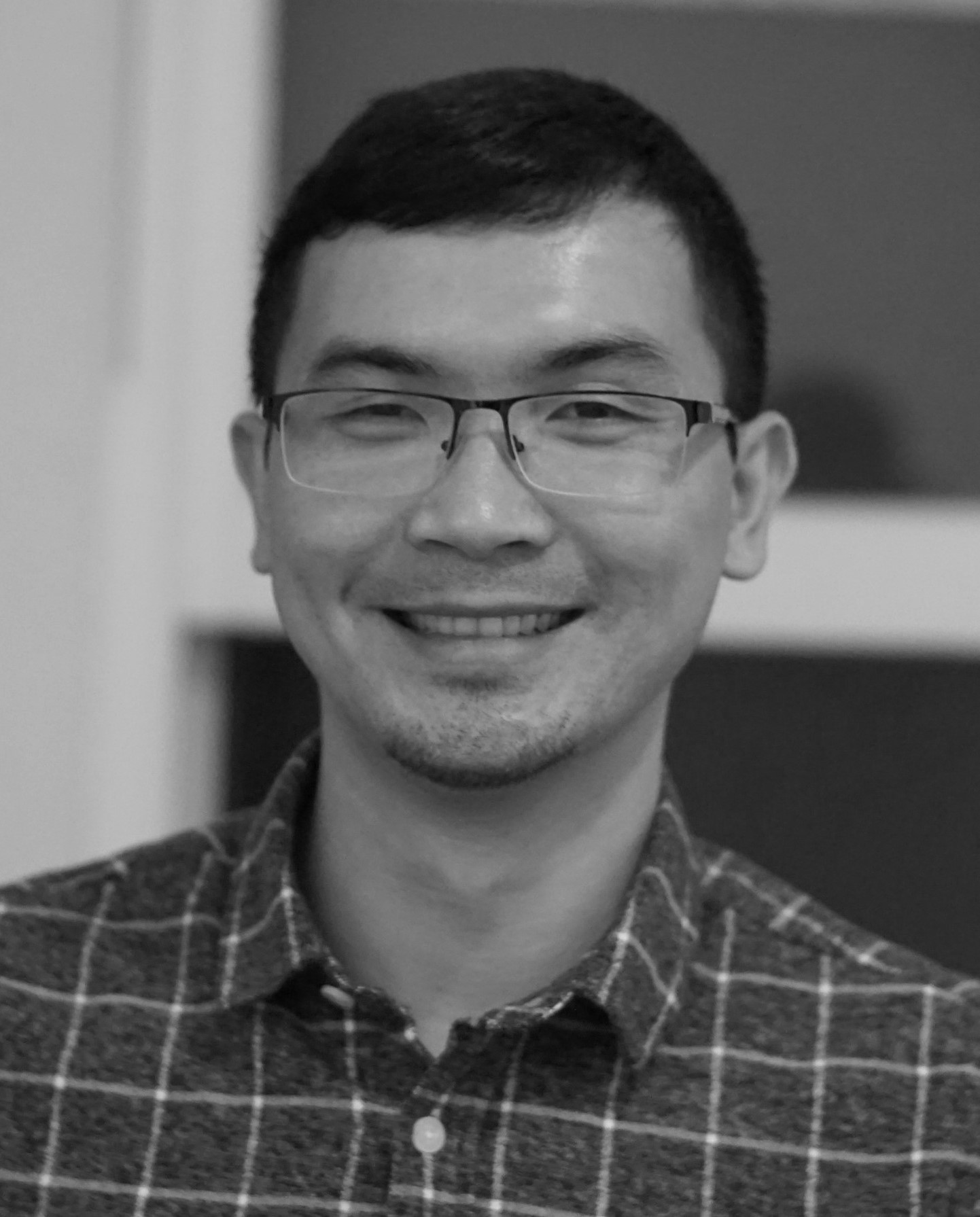}}]{Manh Duong Phung} received the B.Sc. and Ph.D. degrees from Vietnam National University, Hanoi, Vietnam in 2005 and 2015, respectively. 
	
	He is currently a lecturer at University of Technology Sydney and Vietnam National University, Hanoi. He has conducted a number of research projects with industry and international partners such as Eye tracking with NTT Cyber Solution Laboratory, Japan, Telehealth with Mechatronics and Automation Laboratory of National University of Singapore, 3-D hand tracking with Samsung Vietnam Mobile R\& D Center, and Robotics and Automation in Construction with Department of Defense, Australia. His research interests include automation in construction, unmanned aerial vehicles, mobile robot localization and mapping, and optimization.	
\end{IEEEbiography}

\begin{IEEEbiography}[{\includegraphics[width=1in,height=1.25in,clip,keepaspectratio]{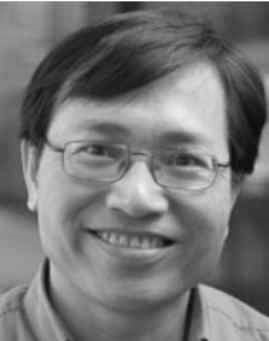}}]{Q. P. Ha} (SM’13) received the B.E. degree from Ho Chi Minh City University of Technology, Ho Chi Minh City, Vietnam, in electrical engineering in 1983 and the Ph.D. degrees from Moscow Power Engineering Institute, Moscow, Russia, in complex systems and control in 1993, and the University of Tasmania, Australia, in intelligent systems, in 1997.
	
	He is currently an Associate Professor with the School of Electrical and Data Engineering of the Faculty of Engineering and Information Technology, University of Technology, Sydney, Australia. His research interests include automation, robotics, and control systems. Dr. Ha has been on the Board of Directors of the International Association
	of Automation and Robotics in Construction since 2007. He was Conference Chair/Co-Chair of several international conferences on automation and intelligent systems. 
	
	He has been on the editorial board of the IEEE Transactions on Automation Science and Engineering (2009–2013), Automation in Construction, Robotica, Electronics, and some others. He was the recipient of a number of best paper awards from the IEEE, IAARC, and Engineers Australia, including the Sir George Julius Medal in 2015.	
\end{IEEEbiography}






\balance

\EOD
\end{document}